\documentclass[preprint,journal]{vgtc}       % preprint (journal style)
\ifpdf%                                % if we use pdflatex
  \pdfoutput=1\relax                   % create PDFs from pdfLaTeX
  \usepackage{graphicx}                % allow us to embed graphics files
  \DeclareGraphicsExtensions{.pdf,.png,.jpg,.jpeg} % for pdflatex we expect .pdf, .png, or .jpg files
\else%                                 % else we use pure latex
  \usepackage{graphicx}                % allow us to embed graphics files
  \DeclareGraphicsExtensions{.eps}     % for pure latex we expect eps files
\fi%

\graphicspath{{figures/}{pictures/}{images/}{./}} % where to search for the images

\usepackage{microtype}                 % use micro-typography (slightly more compact, better to read)
\PassOptionsToPackage{warn}{textcomp}  % to address font issues with \textrightarrow
\usepackage{textcomp}                  % use better special symbols
\usepackage{mathptmx}                  % use matching math font
\usepackage{times}                     % we use Times as the main font
\usepackage{cite}                      % needed to automatically sort the references
\usepackage{tabu}                      % only used for the table example
\usepackage{booktabs}                  % only used for the table example

\usepackage{comment}
\usepackage{multirow}
\usepackage{tabu}                      % only used for the table example
\usepackage{booktabs}                  % only used for the table example
\usepackage{algorithm}
\usepackage{algorithmic}
\preprinttext{Accepted to IEEE VIS 2022. To appear in IEEE Transactions on Visualization and Computer Graphics.}

\onlineid{0}

\vgtccategory{Research}
\vgtcpapertype{please specify}

\title{D-BIAS: A Causality-Based Human-in-the-Loop System \\ for Tackling Algorithmic Bias}

\author{Bhavya Ghai and Klaus Mueller, \textit{Senior Member, IEEE}}

\authorfooter{
\item Bhavya Ghai and Klaus Mueller are affiliated to the Computer Science department, Stony Brook University. E-mail: \{bghai, mueller\}@cs.stonybrook.edu.
}

\shortauthortitle{Ghai and Mueller}
\abstract{
With the rise of AI, algorithms have become better at learning underlying patterns from the training data including ingrained social biases based on gender, race, etc.
Deployment of such algorithms to domains such as hiring, healthcare, law enforcement, etc. has raised serious concerns about fairness, accountability, trust and interpretability in machine learning algorithms.
To alleviate this problem, we propose D-BIAS, a visual interactive tool that embodies human-in-the-loop AI approach for auditing and mitigating social biases from tabular datasets.
It uses a graphical causal model to represent causal relationships among different features in the dataset and as a medium to inject domain knowledge.
A user can 
detect the presence of bias against a group, say females, or a subgroup, say black females, by identifying unfair causal relationships in the causal network and using an array of fairness metrics.
Thereafter, the user can mitigate bias by refining the causal model and acting on the unfair causal edges. For each interaction, say weakening/deleting a biased causal edge, the system uses a novel method to simulate a new (debiased) dataset based on the current causal model while ensuring a minimal change from the original dataset.  
Users can visually assess the impact of their interactions on different fairness metrics, utility metrics, data distortion, and the underlying data distribution. Once satisfied, %they can use the data simulated with the debiased causal model for any downstream application.
they can download the debiased dataset and use it for any downstream application for fairer predictions.
We evaluate D-BIAS by conducting experiments on 3 datasets and also a formal user study. We found that D-BIAS helps reduce bias significantly compared to the baseline debiasing approach across different fairness metrics while incurring little data distortion and a small loss in utility. 
Moreover, our human-in-the-loop based approach significantly outperforms an automated approach on trust, interpretability and accountability.
} % end of abstract


\CCScatlist{ % not used in journal version
 \CCScat{K.6.1}{Management of Computing and Information Systems}%
{Project and People Management}{Life Cycle};
 \CCScat{K.7.m}{The Computing Profession}{Miscellaneous}{Ethics}
}

\teaser{
  \centering
  \includegraphics[width=\linewidth]{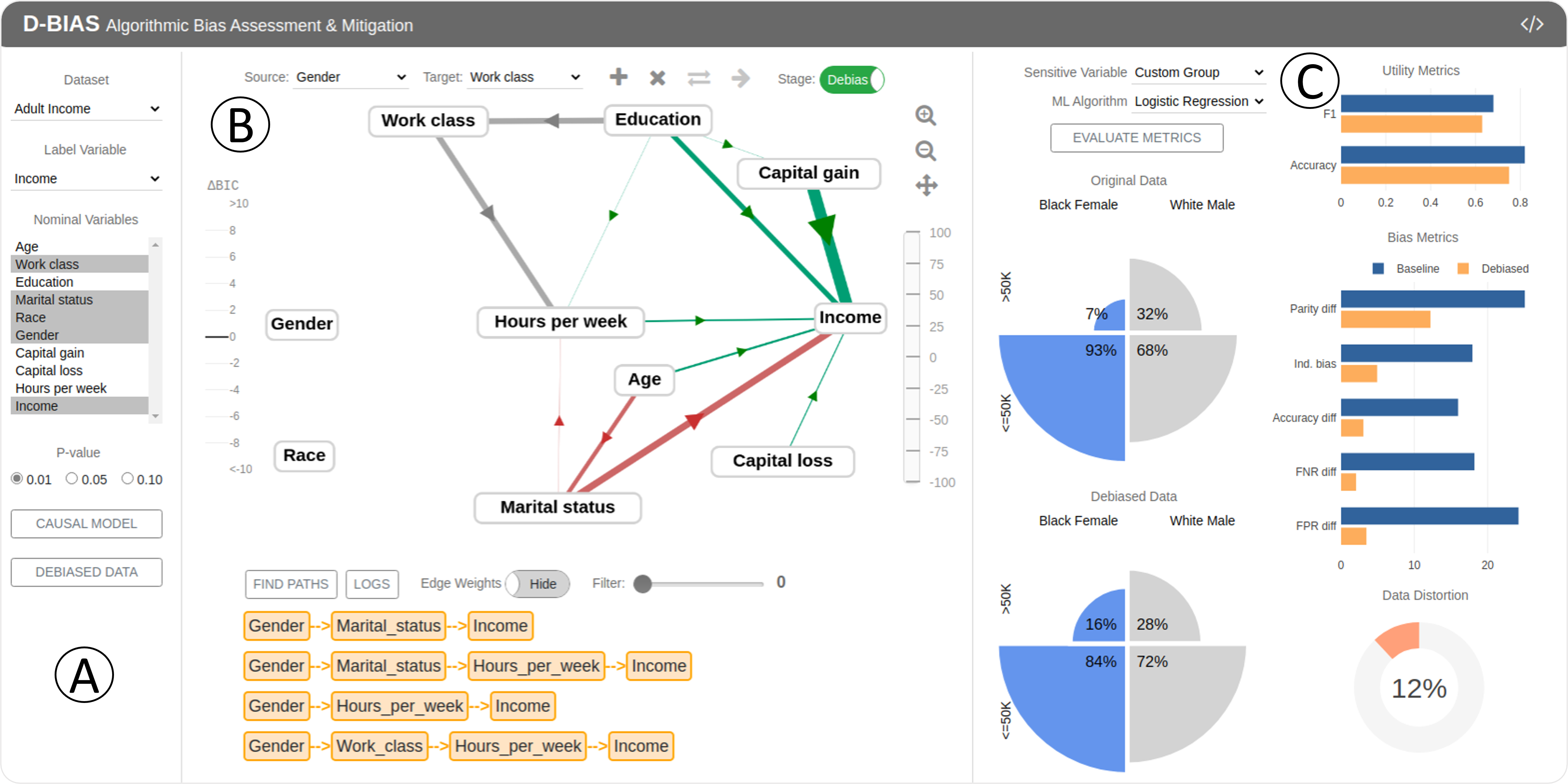}
  \caption{The visual interface of our D-BIAS tool. (A) The Generator panel: used to create the causal network and download the debiased dataset (B) The Causal Network view: shows the causal relations between the attributes of the data, allows user to inject their prior in the system (C) The Evaluation panel: used to choose the sensitive variable, the ML model and displays different evaluation metrics. }  
 \label{fig:teaser}
}

\vgtcinsertpkg

\begin{document}


\firstsection{Introduction}

\maketitle

When computer systems discriminate based on an individual's inherent characteristic such as gender, or acquired traits such as nationality, which are protected classes under law and are irrelevant to the decision making process,
it constitutes algorithmic bias. A simple way to deal with this problem can be to remove the sensitive attribute such as race before training the machine learning (ML) model. However, algorithmic bias can still persist via proxy variables such as zipcode that are correlated with the sensitive attribute. Recent years have seen a huge surge in research papers that deal with this problem. These papers have largely focused on pure algorithmic means to detect and remove bias at different stages of the ML pipeline.
However, fairness is contextual and thus cannot be achieved using fully automated methods \cite{wachter2021fairness}. Moreover, existing techniques are largely black boxes, offering only limited insight on the proxy variables and how bias is mitigated. Finally, they are also limited in providing capabilities that allow users to actively steer and control the debiasing process. Given this limited transparency and human control, accountability and trust become a major concern.

To address these needs, we hypothesize that a human-in-the-loop (HITL) approach is the way forward. A human expert can determine what fairness means in a given context. Such domain knowledge can be incorporated effectively via the HITL approach and hence improve perceived fairness. Introducing a human into the loop will only be effective when a person can understand the underlying state of the system and provide useful feedback. Hence, this approach is naturally inclined towards interpretability. 
On the trust aspect, people are more likely to trust a system if they can tinker with it, even if it meant making it perform imperfectly \cite{dietvorst2016overcoming}.
Human interaction is a core part of the HITL approach, so it might instill more trust.
Lastly, this approach should also foster accountability as the human has a much bigger role to play, which can significantly impact the results.

\begin{figure}[tb]
 \centering 
  \setlength{\belowcaptionskip}{-4pt}
 \setlength{\abovecaptionskip}{-1pt}
 \includegraphics[width=0.9\columnwidth]{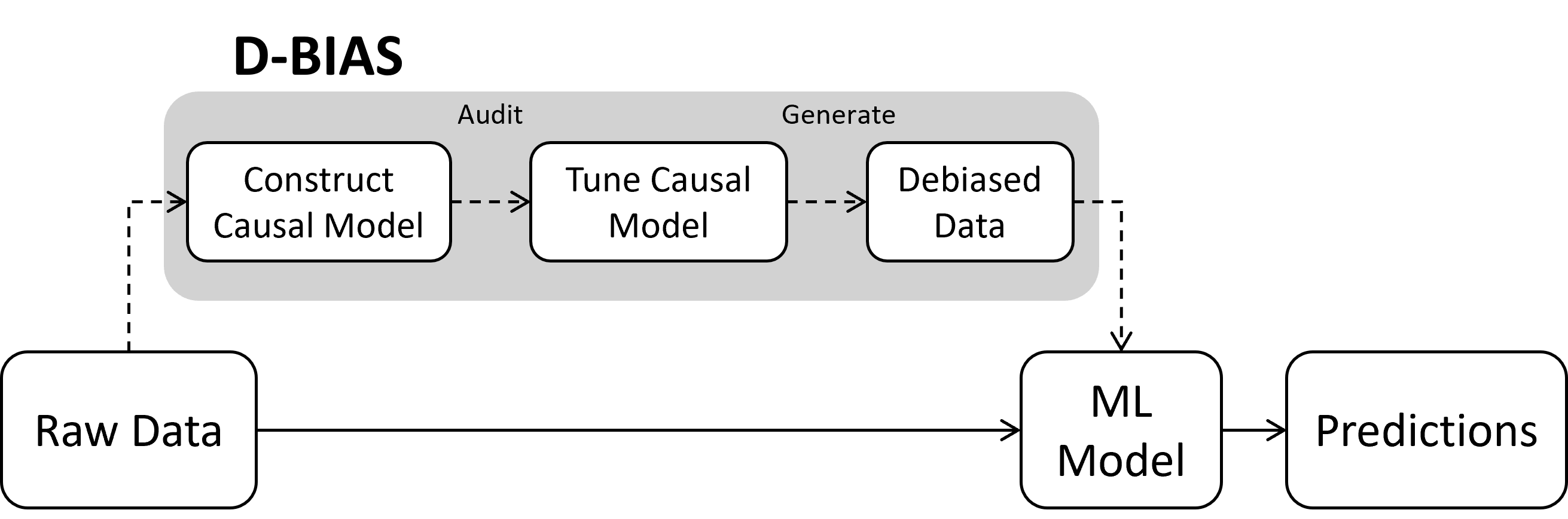}
 \caption{This figure shows how D-BIAS fits into the typical ML pipeline. D-BIAS allows users to act on the raw data and returns its debiased version which can be fed into a ML model for fairer predictions.}
 \label{fig:stages}
 %\vspace{-10}
\end{figure}

We present D-BIAS, a visual interactive tool
%bias/fairness management system 
that embodies a HITL approach for bias identification and mitigation. Given a tabular dataset with meaningful column names as input, D-BIAS assists users in auditing the data for different biases and then helps generate its debiased version (see Fig. \ref{fig:stages}). It uses a graphical causal model as a medium for users to visualize the causal structures inherent in the data and  to inject their domain knowledge. We have made use of causal models since discrimination is inherently causal, and causal models can also be easy to interpret
%also have a natural inclination towards interpretability 
\cite{zhang2017anti, silva,hoque2021outcome}. Apart from causal model, D-BIAS also includes multiple statistical fairness metrics to help identify bias. Users can choose to compare between two groups based on a single variable say gender (Male, Female) or a combination of attributes say race and gender (Black Females, White Males). Thereafter, they can inject their domain knowledge by acting on the edges of the causal network, for instance by deleting or weakening biased causal edges. Since the causal model encodes the data-generating mechanism, any user intervention modifies that process. Following each change, the system generates a new dataset based on the current causal model while keeping track and visualizing the impact of the user interventions on utility, data distortion and various fairness metrics. 
%The final causal model is then used to simulate a new dataset which serves as the debiased dataset. 
Users can interact with the system until they are satisfied with the outcome and then download the debiased dataset for use in any downstream ML application to achieve fairer predictions.
%It allows users to recognize where bias in a dataset occurs and how it is propagated. 
%The visual interface of D-BIAS is backed by causal models which are intuitive to interpret. Having identified the sources of bias. 
%Our system allows users to make the bias-mitigating changes directly in the visual interface. 
%For each change, users can see the change's impact on fairness, data distortion and other utility metrics. 
%D-BIAS affords a very visible and user-driven process that has high potential to increase accountability and trust in the ADM application that uses the debiased data. 
%provides a . Since a human is supervising the process, it brings more accountability and trust in the system.  
The major contributions of our work are:

\begin{itemize}
%\vspace{-0.7em}
\setlength{\itemsep}{-3pt}
  %\item We propose a novel human in the loop method to mitigate biases in tabular datasets using causal models.
  \item A novel human-in-the-loop method to debias tabular datasets.
  \item An end-to-end visual interactive tool for algorithmic bias identification and mitigation.
  %and highlight its potential advantages.
  %\item Our method is embodied in an interactive visual tool which empowers experts and non-experts to understand and deal with bias.
  %\item We demonstrate the effectiveness of our approach using a synthetic and 3 real world datasets.
  \item A demonstration of the effectiveness of our tool in reducing bias using three datasets.
  %\item Using a user study, we evaluate the usability of our tool and compare it against the state of the art on measures like Trust, Interpretability and Accountability.
  \item A user study to evaluate our tool on human-centric measures like usability, trust, interpretability, accountability, etc. 
\end{itemize}

%In our paper, Section 2 presents related work. Section 3 presents our methodology, Section presents our tool, Section 5 presents a case study, Section 6 presents our user study, and Section 7 presents a discussion.  

%The source code for this project can be accessed at \url{bit.ly/3gRGJgA}. It will be made publicly available upon acceptance for easy reproducibility.
%The source code for this project can be found in the supplementary material. It will be made publicly available upon acceptance for easy reproducibility.

%\section{Problem Statement}
%We consider a supervised learning setting where we are given a tabular dataset consisting of a set of input attributes along with their corresponding output labels. Here, each input attribute can be a numeric or categorical variable and the output label will be a binary categorical variable. For example, we might receive data on the attributes of prospective university students and the output variable is whether they got admission. The user can choose to compare between two groups based on a single variable say gender (Male, Female) or a combination of attributes say race and gender (Black Females, White Males). The task is to identify if there is disparity between the two chosen groups. For example, if there is sexism inherent in the dataset which might be amplified by the ML model trained over it. Thereafter, our objective is to make minimal changes to the dataset (attributes and labels) such that the disparity is minimized while the utility (predictive power) is preserved. We aim to achieve both these goals in a human-centered fashion and finally return the debiased dataset.   

\section{Background and Related Work}

\subsection{Bias Identification}
\label{sec:identification}
The existing literature on bias identification mostly revolves around different fairness metrics. Numerous fairness metrics have been proposed which capture different facets of fairness, such as group fairness, individual fairness, counterfactual fairness, etc. 
%\cite{dwork2012fairness,dwork2018group,aif360, bechavod2017penalizing, kate,kusner2017counterfactual,arvindTalk}.
\cite{dwork2018group,aif360, bechavod2017penalizing, kusner2017counterfactual,arvindTalk}.
%For eg., Group fairness imply that members of one subgroup should receive similar proportion of positive/negative outcomes as other subgroups\cite{bechavod2017penalizing, kate}, Individual fairness imply that similar individuals should be treated similarly \cite{dwork2012fairness, dwork2018group}, etc. 
Another way to classify fairness metrics can be on the level they operate on. For eg., dataset based metrics are solely computed using the dataset and are independent of any ML model, such as statistical parity difference. On the other hand, classifier based metrics are computed over the predictions of a trained ML model, such as false negative rate difference. 
So far, there is not a single best fairness metric. Moreover, some of the fairness metrics can be mutually incompatible, i.e., it is impossible to optimize different metrics simultaneously\cite{kleinberg2016inherent}. 
%Hence, it is advisable to consider multiple fairness metrics to present a more comprehensive picture. 
%Due to computational and space constraints, its difficult to include all fairness metrics in the visual tool. Hence 
%In line with existing visual tools \cite{silva, fairvis}, we have used as set of popular fairness metrics to quantify bias (see \autoref{fig:teaser} (E)). 
In line with existing visual tools \cite{silva, aif360}, our tool also uses a diverse set of fairness metrics to present a more comprehensive picture.

Many fairness metrics solely focus on the aggregate relationship between the sensitive attribute and the output variable.  
%It is possible that some of the underlying factors causing biased outcomes might be legitimate/fair. 
This can lead to misleading conclusions as the aggregate trend might disappear or reverse when accounting for other relevant factors. A prime example of this phenomenon, also known as Simpson's paradox \cite{Pearl2009}, is the Berkeley Graduate Admission dataset \cite{bickel1975sex}. There it appeared as if the admission process was biased against women since the overall admit rate for men (44\%) was significantly higher than for women (30\%) \cite{barocas2017fairness}. However, this correlation/association did not account for the fact that women typically applied for more competitive departments than men. After correcting for this factor, it was found that the admission process had a small but statistically significant bias \textit{in favor of} women \cite{bickel1975sex}.
%examining the data at the department level found that 4 of the 6 largest departments had a higher admit rate for women. 
Causal models can be an effective tool for dealing with such a situation as they can decipher the different intermediate factors (indirect effects) along with their respective contributions behind an aggregate trend. 
%model direct as well as indirect effects.  
%Causal models were able to conveniently express this dual influence of gender on admit rate, namely, as a direct cause but also indirectly through the mediating variable \textit{department}. 
Hence, our tool also employs causal model for bias identification.

\begin{figure*}
 \centering 
 \includegraphics[width=2\columnwidth]{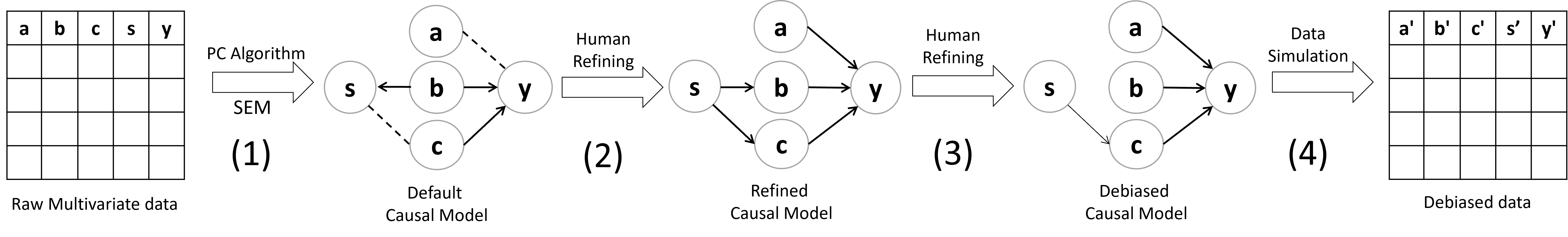}
  \setlength{\belowcaptionskip}{-8pt}
\setlength{\abovecaptionskip}{-4pt}
 \caption{The workflow of D-BIAS has 4 parts: (1) The default causal model is inferred using the PC algorithm and SEM. Here, dotted edges represent undirected edges (2) Refine stage - the default causal model is refined by directing undirected edges, reversing edges, etc. (3) Debiasing stage - the causal model is debiased by deleting/weakening biased edges (4) The debiased causal model is used to simulate the debiased dataset.}  
 \label{fig:workflow}
\end{figure*}

\subsection{Bias Mitigation}
\label{sec:mitigation}
The existing literature on algorithmic bias mitigation can be broadly categorized into the three different stages in which they operate within the ML pipeline, namely pre-processing, in-processing and post-processing. In the pre-processing stage, the dataset is modified such that the bias with respect to an attribute or set of attributes is reduced or eliminated \cite{zemel2013learning, kamiran2009classifying, hajian2013dataModification, nipsIBM, rajabi2021tabfairgan}. This can be achieved by either modifying the output label \cite{kamiran2012, kamiran2009classifying} or by modifying the input attributes \cite{nipsIBM, zemel2013learning}. 
%Previous such studies have formulated fairness as an optimization problem. They have tried to minimize distortion, counter discrimination, and preserve utility \cite{nipsIBM, zemel2013learning}. 
In the in-processing stage, the algorithm is designed to take in biased data but still generate unbiased predictions. This can be achieved by tweaking the learning objectives such that accuracy is optimized while discrimination is minimized %\cite{zafar2017fairness,inprocess,kamishima2011}
\cite{inprocess}. Finally, in the post-processing stage, predictions from ML algorithms are modified to counter bias 
%\cite{postProcessing, lohia2018bias}
\cite{lohia2018bias}. 
%Some recent work uses a combination of existing fairness enhancing techniques at different stages of the ML pipeline to achieve more fairness \cite{ghai2022cascaded, feldman2021endtoend}. 
Our work relates closely with the pre-processing stage where we 
%generate the debiased dataset by 
make changes to the input attributes and the output label.  
%\textcolor{blue}{We chose the pre-processing stage since it provides maximum flexibility}. 
%Once the dataset is debiased, it can be then be fed into any machine learning algorithm for more fairer results. 

There is also a growing set of work at the intersection of bias mitigation and causality \cite{zhang2017causal, wu2018discrimination, chiappa2019path}. The general idea is to generate the causal network, modify it, and then simulate debiased data. These approaches fully rely on automated techniques to yield the true causal network, and assume a priori knowledge about fair/unfair causal relationships. 
%They attempt to mitigate bias based on causality based fairness metric by solving an optimization problem.  
Our work draws inspiration from this line of work and presents a more general solution where human domain knowledge is leveraged to refine the causal network and generate the debiased dataset. 
%Unlike previous approaches, our approach allows the user to better understand the impact of causal interventions on a diverse set of fairness metrics and utility metrics. 
%This allows them to take a more informed decision on the best tradeoff between different metrics    

   %There is another line of research which is focused on mitigating bias when protected variable is not available. For the purpose of this work, we assume that protected variable is available.

%\begin{comment}
%visualization community has focused on cognitive biases in visualization. Visual interfaces can play a key %role in bolstering trust & transparency in automated decision making processes. 
%\end{comment}

\subsection{Visual Tools}
Recent years have seen visual tools like \textit{Silva}\cite{silva}, \textit{FairVis}\cite{fairvis}, \textit{FairRankVis}\cite{xie2021fairrankvis}, \textit{DiscriLens}\cite{DiscriLens}, \textit{FairSight}\cite{fairsight}, \textit{WordBias}\cite{ghai2021wordbias}, etc. which are all aimed at tackling algorithmic bias. Although most of these tools are focused on bias identification, a few of them, such as FairSight, also permit debiasing. However, the debiasing strategy used in such tools is fairly basic, like eliminating proxy variable(s). Simple measures like this can lead to high data distortion and can have a high negative impact on data utility. Our work relates more closely with \textit{Silva} which also features a graphical causal model in its interface. Using an empirical study, it showed that users can interpret causal networks and found them helpful in identifying algorithmic bias. However like most other visual tools, Silva is limited to bias identification. Our work advances the state of the art by presenting a tool that supports both bias identification and mitigation, with our debiasing strategy being more nuanced and sensitive toward data distortion. 
\section{Methodology}
The workflow of our system can be understood from Fig. \ref{fig:workflow}. A detailed discussion for each stage follows next.
\begin{comment}
Given raw tabular data as input, D-BIAS outputs its debiased version. The entire process can be understood in four steps as shown in \autoref{fig:workflow}. In step 1, the raw multivariate data is read in and is used to compute a causal model using automated techniques. 
%Here, the user needs to identify the label and nominal attributes (see Fig \ref{fig:teaser}(a)). These are used by the automated causal inference algorithm (see sec \ref{sec:causal} for details) to generate the causal model and for computing evaluation metrics. 
In step 2, the user interacts with the edges of the initial causal model like directing undirected edges to reach to a reliable causal model. In step 3, the user identifies and modifies biased causal relationships. Finally, in step 4, our tool computes a new dataset based on user interactions in the previous step. The user can thereafter click on `Evaluate Metrics' to compute and plot different evaluation metrics. 
If the user is satisfied with the evaluation metrics, they can download the debiased dataset and safely use it for any downstream application. Otherwise, the user may return to the causal network and make further changes. This cycle continues until the user is satisfied. A detailed discussion follows next. 
\end{comment}

\subsection{Generating the Causal Model}
\label{sec:causal}
A causal model generally consists of two parts: causal graph (skeleton) and statistical model. Given a tabular dataset as input, we first infer an initial causal graph (directed acyclic graph) using a popular causal discovery algorithm, namely the PC algorithm  %(named after its inventors \underline{P}eter Spirtes and \underline{C}lark Glymour) 
\cite{Spirtes2000,Colombo2014Order-independentLearning}.
%The PC algorithm starts with a complete undirected graph where each node is an attribute in the dataset. Thereafter, it starts filtering edges based on conditional independence tests. 
%For example, if nodes X and Y are unconditionally independent, then the edge $X \to Y$ is removed.
This algorithm relies on conditional independence tests and a set of orientation rules to yield the causal graph (for details, see appendix B). Each node in the causal graph represents a data attribute and each edge represents a causal relation. For example, a directed edge from X to Y signifies that X is the cause of Y, i.e., a change in X will cause a change in Y.
%Here, the tabular dataset can consist of numerical and categorical attributes; we have used the \textit{Symmetric Conditional Independence Test} which can deal with mixed data types \cite{tsagris2018constraint}. 
%Each node in the causal graph represents a variable in the dataset that can be bifurcated into an exogenous or an endogenous variable. An exogenous variable is an independent variable i.e., it is not affected by any other variable while endogenous variables are determined as a function of their parents.
%After filtering, it uses a set of orientation rules to direct edges. Each directed edge represents a causal relationship from cause to effect. PC algorithm returns a partially directed acyclic graph i.e., a DAG with some undirected edges. 
The PC algorithm provides qualitative information about the presence of causal relationship between variables. To quantitatively estimate the strength of each causal relationship, 
%we have used a specialized case of structural equation modeling (SEM) i.e., Path Analysis \cite{Pearl2000}. 
we use linear structural equation models (SEM) where a node is modeled as a linear function of its parent nodes. 
%(for details, see appendix B.2). 
%with some error term (intercept). 
%For better interpretability, D-BIAS uses linear SEMs i.e., each node is modeled as a \textit{linear} function of its parents. 
%using a linear or logistic regression model.

\begin{equation}
y = \sum_{i}^{parents(y)}{\beta_{i} x_i} \;+ \;\varepsilon 
\end{equation}

In the above equation, variable y is modeled as a linear combination of its parents $x_i$, their regression coefficients ($\beta_i$) and the intercept term ($\varepsilon$). 
%If y is a numerical variable, we use linear regression to model this relationship else we use the multinomial logit model. Here, $\beta_i$ represents the strength of causal relation between $x_i$ and y. 
If y is a numerical variable, we use linear regression else we use the multinomial logit model to estimate the values of $\beta_i$ and $\varepsilon$. Here, $\beta_i$ represents the strength of causal relation between $x_i$ and y. 
We repeat this modeling process for each node with non-zero parent nodes. Such nodes that have at least one edge leading into them are termed as endogenous variables.  
%We repeat this process for all endogenous variables, i.e., all nodes that have at least one edge leading into them. 
Other nodes correspond to exogenous variables or independent variables that have no parent nodes. In \autoref{fig:workflow} (Refined Causal Model), s and a are exogenous variables while b, c and y are endogenous variables. Here, y will be modeled as a function of its parent nodes, i.e., a, b and c. Similarly, b and c will be modeled as function of s.
 After this process, we arrive at a causal graph whose different edges are parameterized using SEM. This constitutes a Structural Causal Model (SCM) or simply causal model (see Fig. \ref{fig:workflow} (1)).  
%Apart from estimating the strength of causal relationships, SEMs also help in refining the causal model by providing statistics on how well a given causal model fits the underlying dataset. This provides a feedback to the user whenever they modify an edge in the causal graph.

%This causal model might have some misdirected or undirected causal edges due to variables that have not been captured or latent variables that can't be numerically assessed. 
This causal model, generated using automated algorithm, might have some undirected/erroneous causal edges due to different factors like sampling bias, missing attributes, etc.   
To achieve a reliable causal model, we have taken a similar approach as Wang and Mueller \cite{wang2017visual} where we leverage user knowledge to refine the causal model via operations like adding, deleting, directing, and reversing causal edges (see Fig. \ref{fig:workflow} (2)). For every operation, the system computes a score (Bayesian Information Criterion (BIC)) of how well the modified causal model fits the underlying dataset. Similar to \cite{wang2017visual}, our system assists the user in refining the causal model by providing the difference in BIC score before and after the change. A negative BIC score suggests a better fit.  
After achieving a reliable causal model, we enter into the debiasing stage where any changes made to the causal model will reflect on the debiased dataset (see Fig. \ref{fig:workflow} (3)).

\subsection{Auditing and Mitigating Social Biases}
From a causal perspective, discrimination can be defined as an unfair causal effect of the sensitive attribute (say, race) on the outcome variable (say, getting a loan)\cite{Pearl2000}. A direct causal path from a sensitive variable to the output variable constitutes disparate treatment while an unfair indirect path via some proxy attribute (say, zipcode) constitutes disparate impact \cite{chiappa2018causal}. 
A direct path is certainly unfair but an indirect path may be fair or unfair (as in the case of Berkeley Admission dataset \cite{Pearl2009}). Our system computes all causal paths 
%from a sensitive attribute to the outcome variable. Thereafter, it is the job of 
and lets the user decide if a causal path is fair or unfair. If a causal path is unfair, the user should identify which constituting causal relationship(s) are unfair and act on them. The user can do this by deleting or weakening such biased causal relationships to reduce/mitigate the impact of the sensitive attribute on the outcome variable. For example, in Fig.\ref{fig:workflow} (Refined Causal Model), s is the sensitive attribute and y is the outcome variable. Here, the user deletes the edge $s \to b$ and weakens the edge $s \to c$ (shown as a thin edge). Once the user has dealt with the biased causal relationships, we achieve what we call the \textit{Debiased Causal Model}.  
%A directed path from a sensitive variable S such as gender to an output variable Y such as, getting a job, constitutes \textit{direct discrimination} or \textit{disparate treatment}. On the other hand, a directed path from S to Y via some other attribute(s) A constitutes \textit{disparate impact}. Here, A is known as proxy variable(s).   
%If there is a direct path, it means the sensitive attribute directly impacted the output attribute (disparate treatment). An indirect path means that the sensitive attribute is impacting the output attribute via proxy variables (disparate impact).
%\textcolor{blue}{Berkeley example}
%\textcolor{blue}{explain more with figure, some paths might be legitimate, other may not}

%\textcolor{blue}{During debiasing stage, addition operation should performed first followed by deletion and modifying edge weight.}

\begin{algorithm}
\caption{Generate Debiased Dataset}
\label{algo:1}
\begin{algorithmic}[1]
\STATE $ D \leftarrow $ Original Dataset
%\STATE // G - Initial Causal Graph with vertices V and edges E
%\STATE $G(V, E) \leftarrow PC\_algorithm(D)$ \hfill // G - Causal Graph
\STATE $G(V, E) \leftarrow $ refined causal model   %\hfill // G - Causal Graph
\STATE $E_a \leftarrow$ set of edges added during the debiasing stage
\STATE $E_m \leftarrow $ subset of E that were deleted/strengthened/weakened during the debiasing stage
%\STATE \textcolor{blue}{What about added edges?}
\STATE
%\STATE // re-learn weights for 
\FOR{\textbf{each} e in $E_a$}
\STATE $n \leftarrow $node pointed by head of e
\STATE retrain linear model for n as a function of its parents 
\ENDFOR
\STATE
\STATE $V_{sim} \leftarrow \emptyset$ \hfill // Attributes that need to be simulated
\FOR{\textbf{each} e in $(E_m \cup E_a)$}
%\IF{e present in E'}
\STATE $n \leftarrow $node pointed by head of e
\STATE $V_{sim} \leftarrow V_{sim} + n + all\_descendent\_nodes(n)$
%\ENDIF
\ENDFOR
\STATE $V_{sim} \leftarrow remove\_duplicates(V_{sim})$
\STATE
\STATE $D_{deb} \leftarrow \emptyset $ \hfill // Debiased Dataset
\FOR{\textbf{each} v in topological\_sort(V)}
\IF{v present in $V_{sim}$}
\STATE $D_{deb}[v] \leftarrow $ generate values based on \autoref{eq:gen}
\ELSE
\STATE $D_{deb}[v] \leftarrow D[v]$
\ENDIF
\ENDFOR
\STATE
\STATE // Rescale values for the simulated attributes
\FOR{\textbf{each} v in $V_{sim}$} 
\IF{v is a categorical variable}
\STATE $D_{deb}[v] \leftarrow $ rescale values based on Algorithm 2
\ELSE 
\STATE // v is a numeric variable 
\STATE $\mu, \sigma^2 \leftarrow mean(D[v]), variance(D[v])$
\STATE $\mu_{deb}, \sigma_{deb}^2 \leftarrow mean(D_{deb}[v]), variance(D_{deb}[v])$
\STATE $D_{deb}[v] \leftarrow \mu + (D_{deb}[v] - \mu_{deb})/\sigma_{deb}*\sigma$ 
%standardize $D\_deb[v]$ values to have mean $\mu$ and variance $\sigma$
\ENDIF
\ENDFOR
\STATE \textbf{Result} $D_{deb}$
\end{algorithmic}
\end{algorithm}

%\vspace{-1em}
\subsection{Generating the Debiased Dataset}
%Given the debiased causal model, we will \textit{simulate} a dataset which embodies the final causal network while having 
%Once the user has refined the causal model and dealt with the biased causal relationships,
%included their prior in the causal network, 
We simulate the debiased dataset based on the debiased causal model (see Fig. \ref{fig:workflow} (4)). 
%This simulated dataset inherits all the patterns which are present in the causal model. 
The idea is that if the user weakens/removes biased edges from the causal network, then the simulated dataset might also contain less biases. The standard way to simulate a dataset based on a causal model is to generate random numbers for the exogenous variables \cite{sofrygin2017simcausal}. Thereafter, each endogenous variable is simulated as a function of its parent nodes (variables) in the causal network. 
In this work, we have adapted this procedure to suit our needs, i.e., simulating a fair dataset while having minimum distortion from the original dataset. 
%(see Algorithm \ref{algo:1}).
%In this work, we have adapted this procedure to use the training data for simulation. It enables us to make meaningful comparisons of the fairness outcomes achieved by our various interventions. 

%to suit our needs i.e., simulating a fair dataset while having minimum distortion.

%\vspace{-5pt}

Our approach to generate the debiased dataset, as illustrated in Algorithm \ref{algo:1}, can be broken down into 4 steps.
%\textbf{Data debiasing algorithm.}  our approach has 4 steps. 
At step 1 (lines 6--9), we focus on the set of edges added during the debiasing stage ($E_{a}$). We retrain regression models corresponding to each of the target nodes of $E_{a}$. This will update the weights (regression coefficients ($\beta_i$)) for all edges leading into any target node of $E_{a}$. At step 2 (lines 11--16), we identify the set of nodes (attributes) that need to be simulated ($V_{sim}$). Unlike the standard procedure, we only simulate selective nodes that are directly/indirectly impacted by the user's interaction to minimize distortion. This set includes the target nodes of all edges that the user has interacted with along with their descendent nodes. For example, in \autoref{fig:workflow}(3), the user deletes the edge $s \to b$ and weakens the edge $s \to c$. So, we will only simulate variables $b$, $c$ and $y$. At step 3 (lines 18--25), we actually simulate all nodes that are a part of $V_{sim}$ using \autoref{eq:gen}. All other nodes are left untouched and their values are simply copied from the original dataset into the debiased dataset. 
It should be noted that all parent nodes must be simulated before their child node as the values for a node are computed using their parent nodes. So, we simulate all endogenous variables in a topological order.
%It should be noted that an endogenous variable might have other endogenous variable(s) as its parent node(s). So, we simulate endogenous variables in a topological order so that all parent nodes have been simulated before the child node is. 
For eg., in \autoref{fig:workflow}(4), nodes \textit{b} and \textit{c} will be simulated before node \textit{y}.  
%It results in a dataset where variables previously affected by bias now have values that better adhere to the user's view of fairness.

\vspace{-7pt}
%\vspace{-1em}
\begin{equation}
\label{eq:gen}
%y_i = \sum_{j}^{parents(i)}{\alpha_{ij}\beta_{ij} x_{j}} + \varepsilon_{i} +  \sum_{j}^{parents(i)}{(\beta_{ij}-\alpha_{ij}\beta_{ij}) r_{j}}
y = \sum_{i}^{parents(y)}{\alpha_{i}\beta_{i} x_{i}} \;+\;\; \varepsilon  \;\;+\; \sum_{i}^{parents(y)}{(1-\alpha_{i})\beta_{i}r_{i}}
\end{equation}
\vspace{-7pt}

%\vspace{-1em}
In the above equation, node y is simulated as a sum of 3 terms. The first term is the weighted linear combination of parent nodes. Here, $\alpha_i$ is the scaling factor that has a default value of 1 and can range between [0, 2] as determined by user interaction. 
%Setting $\alpha_i$$>$1 amplifies the edge, while $\alpha_i$$<1$ weakens it. 
Strengthening an edge, sets $\alpha_i>$1; weakening an edge, sets $\alpha_i<$1. For example, if the user weakens the edge between node $x_i$ and y by -35\%, then $\alpha_i$=0.65, strengthening it by +35\% will set $\alpha_i$=1.35, and deleting it will set $\alpha_i$=0. The second term is the intercept that was computed when the regression model for y was last trained. The third term adds randomness in proportion to the degree to which the user has altered an incoming edge. Here, $r_{i}$ is a normal random variable that has a similar distribution as $x_{i}$ ($r_{i} \sim \mathcal{N}({\mu_{x_{i}}, \sigma_{x_{i}}^{2}})$). This term adds fairness as it is random and alleviates distortion for y.
%as it follows the same distribution. 

\begin{algorithm}[t]
\caption{Rescale Categorical Variable v}
\begin{algorithmic}[1]
\STATE D[v] $\leftarrow$ categorical variable v in the original dataset
\STATE prob\_mat $\leftarrow$ probability matrix for v computed using Eq. 2 
\STATE
\STATE lr $\leftarrow $ 0.1  \hfill // learning rate
\STATE iterations $\leftarrow $ 0
\LOOP
    %\STATE // each unique value along with its probability
    \STATE // DPD: Discrete Probability Distribution 
    \STATE $dist_{ori} \leftarrow$ DPD( D[v] )
    \STATE $dist_{deb} \leftarrow$ DPD( argmax(prob\_mat) )
    \STATE diff $\leftarrow \sum{\left\|(dist_{ori} - dist_{deb})/dist_{deb}\right\|}$
    \STATE scale\_factor $\leftarrow 1 + lr*\sum{(dist_{ori} - dist_{deb})/dist_{deb})}$
    \STATE prob\_mat $\leftarrow$ scale\_factor * prob\_mat
    \STATE $dist_{deb} \leftarrow$ DPD( argmax(prob\_mat) )
    \STATE new\_diff $\leftarrow \sum{\left\|(dist_{ori} - dist_{deb})/dist_{deb}\right\|}$
    \IF{new\_diff $>$ diff or iterations $>$ 50}
    \STATE break
    \ENDIF
    %\STATE diff = new\_diff
    \STATE iterations $\leftarrow$ iterations + 1
\ENDLOOP
\STATE $D_{deb}[v] \leftarrow$ argmax(prob\_mat)
\STATE \textbf{Result} $D_{deb}[v]$

\begin{comment}
def scale_prob_matrix(pmat, col_ori):
    learning_rate = 0.1
    iterations = 0
    while True:
        m1 = col_ori.value_counts(normalize=True)
        m2 = pmat.idxmax(axis=1).value_counts(normalize=True)
        gap = np.sum(np.abs((m1-m2)/m2))
        # print(gap)
        scale_fac = 1 + learning_rate*((m1-m2)/m2)
        pmat = scale_fac * pmat
        m2 = pmat.idxmax(axis=1).value_counts(normalize=True)
        new_gap = np.sum(np.abs((m1-m2)/m2))
        #print(round(new_gap - gap,10))
        if new_gap > gap or iterations > 50:
            break
        gap = new_gap
        iterations = iterations + 1
    tmp_data.idxmax(axis=1).tolist()
    return pmat
\end{comment}
\end{algorithmic}
\label{algo:2}
\end{algorithm}

\begin{figure}[t]
 \centering 
 \includegraphics[width=0.75\columnwidth]{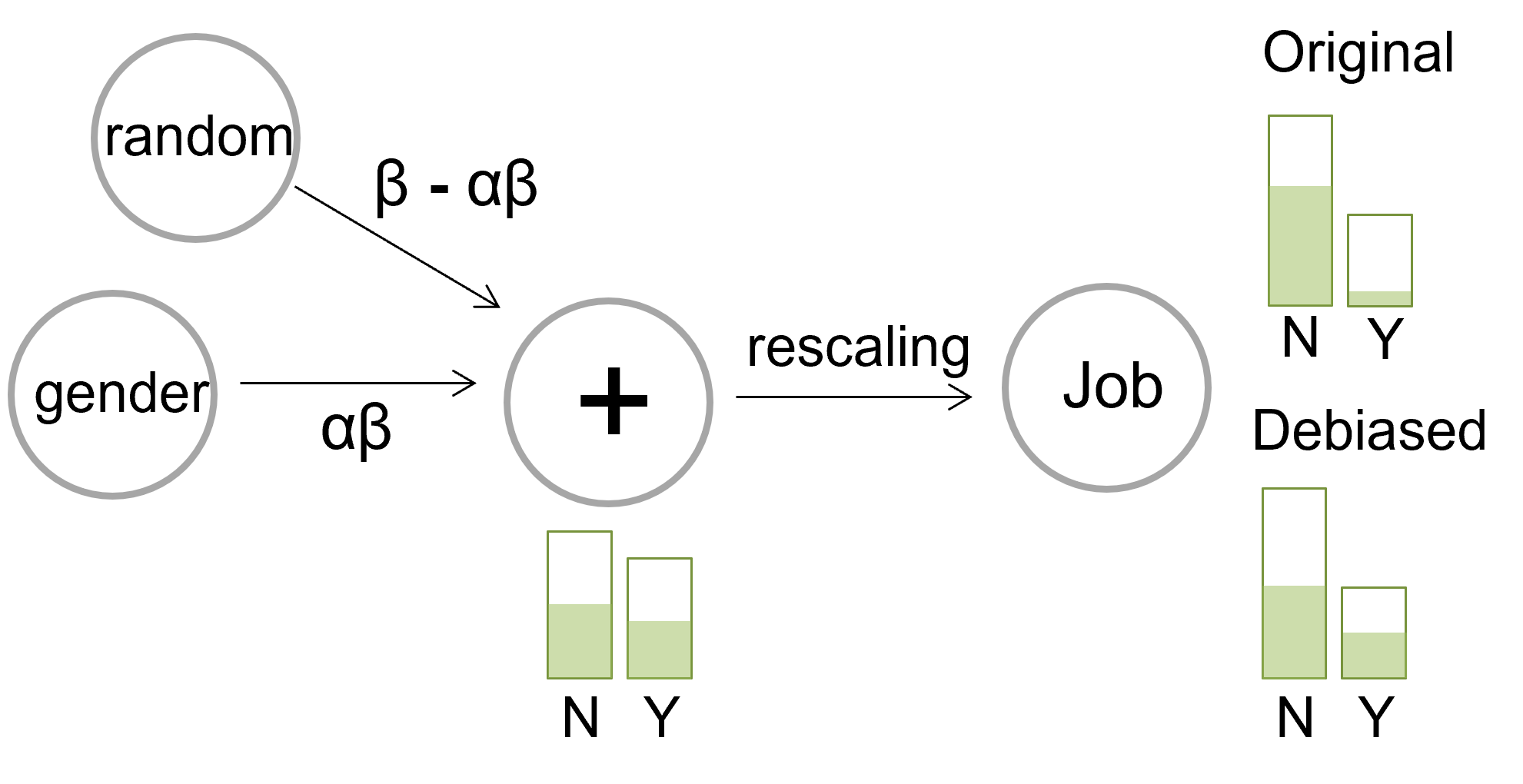}
  \setlength{\belowcaptionskip}{-4pt}
  \setlength{\abovecaptionskip}{-2pt}
 \caption{Illustration of the data debiasing process using a toy example where the node Job is caused by a single variable, namely gender. Green color marks the proportion of females who got the job (Y) or not (N).}  
 \label{fig:deb_eqn}
%\vspace{-6}
\end{figure}

Fig. \ref{fig:deb_eqn} illustrates the case where the node \textit{Job} has a single parent node (\textit{Gender}). Let's say that the user chooses to delete this edge ($\alpha$=0). Going the conventional route (without the third term), the attribute \textit{Job} will be reduced to a constant value (the intercept ($\varepsilon$)) which is undesirable.
%(the second term in \autoref{eq:gen}, the intercept computed for the causal model). 
Adding the (third) random term generates the \textit{Job} distribution below the `+' node which is far more balanced (fair) in terms of \textit{Gender} than the `Original' distribution on the top right. However,
%due to modeling error and added randomness, 
the number of people getting the job (or not) is distorted compared the `Original' distribution. 

This is corrected in Step 4 (lines 27--37) in Algorithm \ref{algo:1}, where we rescale each simulated variable so that its distribution is close to its original distribution. For numerical variable(s), we simply standardize values to their original mean and standard deviation. For categorical variable(s), the model returns a set of probability scores corresponding to each possible output label for each data point. 
%We modulate such a probability matrix to match the original distribution. As described in Algorithm \ref{algo:2}, we iteratively scale the probability matrix so that it inches toward the original distribution. 
We iteratively scale such a probability matrix so that the resulting debiased distribution inches toward the original distribution (see Algorithm \ref{algo:2}).
We continue this process until a fixed number of iterations or when the difference between original and debiased distribution starts increasing. In Fig. \ref{fig:deb_eqn}, we observe that the resulting `Debiased' distribution matches the `Original' distribution in terms of \textit{Job} allocation, while maintaining gender parity. 
%in the hired group achieved by the randomization in Step 3. 
%This entire process generates a new (debiased) dataset which inherits all the characteristics of the debiased causal model while having minimum distortion from the original dataset.  
%In such a case, the random term plays a dominant role and ensures that the original distribution of \textit{Job} is preserved. 
%Going the standard way to generate debiased data might yield a more fairer dataset but it might be farther from reality i.e. it might be highly distorted compared to the original dataset.
%mightn't have much resemblance with the original dataset.  
%The second term is the intercept which was computed while generating the causal model.
It should be noted that simulating each attribute adds a corresponding modeling error to the process. This modeling error is typically small but it can potentially overpower the impact of the user's intervention, especially when a user makes a small change, say weakening an edge by 5\%. 
In such a case, the results may not be in strict accordance with the user’s expectations.
%In such a case, the effect of a user intervention might not be in accordance with its desired effect. 
%In such cases, the user's intervention might not result in results might not be in accordance with the user's expectations. 

%For example, if a user deletes an edge between the nodes X and Y, then Y and all its children nodes will be simulated. Similarly, we calculate all the nodes impacted by all user interactions and compute their union. All nodes in this union set will only be modified.  
%Given the list of edges added/deleted/diluted/strengthened by the user, we compute the union of all nodes impacted by each edge. 

%First, we keep the exogenous variables unchanged i.e. we use the exact same values from the original dataset. Second, we  If a user makes some change to an edge connecting nodes s and t, then the nodes impacted by this change will be t and all of its children.  All other nodes will be kept constant.   \textcolor{blue}{how I tried to reduce distortion by only simulating some columns}
%\subsection{How to Use?} 
%Direct all edges first. We can't allow for deleting undirected edges because we don't know either source or target node of the edge should be simulated. So first direct all edges. Explain why? Then impose your prior... .... 

\subsection{Evaluation Metrics}
\label{sec:eval_metrics}
%Once the debiased dataset is generated, it can be evaluated using different metrics that operate at the dataset level, and another set of metrics that operate over the ML model's prediction that is trained using the debiased dataset.  
Once the debiased dataset is generated, it can be evaluated using different metrics that operate at the dataset level and the classifier level. For the second case, the debiased dataset is used to train a ML model chosen by the user, and a set of metrics are computed over the model's predictions. Here, the idea is to evaluate the downstream effects of debiasing.
%and the user can evaluate it using a variety of metrics. Moreover, 
%The user can choose from one of the many ML models to evaluate the impact of debiasing on model's prediction. 
%Here, the premise is that a model trained over the debiased dataset should yield more fairer predictions. 
%the user can select one of several ML models and algorithms to evaluate the gains in fairness via a variety of metrics. 
All the evaluation metrics can be grouped into three broad categories, namely utility, fairness and distortion. It should be noted that there might be a trade-off among the three categories. For eg., reducing bias might cause high data distortion or lower utility. For comparison, we have used a baseline debiasing strategy which just removes the sensitive attribute(s) from the dataset.

\textbf{Fairness.} Our tool presents a diverse set of 5 popular fairness metrics, namely statistical parity difference (Parity diff), individual fairness (Ind. bias), accuracy difference (Accuracy diff), false negative rate difference (FNR diff) and false positive rate difference (FPR diff) \cite{aif360}. Two of these metrics operate at the dataset level (Parity diff, Individual bias) and the rest operate on the classifier’s predictions.
%while the others operate on the user-specified classifier's predictions. 
%There are multiple fairness metrics defined in the literature \cite{aif360}. 
Here, Ind. bias is defined as the mean percentage of a data point's k-nearest neighbors that have a different output label. A lower value for Ind. bias is desirable as it means that similar data points have similar output labels. For the 4 other fairness metrics, we compute some statistic for the two groups say males and females, and then report their absolute difference. This statistic can be ML model dependent, such as accuracy,
%(for Accuracy diff metric)
or model independent, such as the likelihood of getting a positive output label.
% (for Parity diff metric)
Lower values for such metrics suggests more equality between groups. For computing model based metrics, we omit the sensitive attributes(s) and perform 3-fold cross validation using the user-specified ML model with 50:50 train test split ration, and then report the mean absolute difference between groups across the 3 folds.

\textbf{Utility.} The utility of ML models is typically measured using metrics like accuracy, F1 score, etc. 
%In our context, we are interested in measuring the utility of a given ML model when the underlying training dataset is replaced by its debiased version. 
In our context, we are interested in measuring the utility of a ML model when it is trained using the debiased dataset instead of the original dataset.
%a dataset can be attributed to the usefulness of the machine learning algorithm trained over it to predict the original output variable.
To compute the utility for the original dataset, we perform 3-fold cross validation using the user-specified ML model and report the mean accuracy and F1 score. For the debiased dataset, we follow a similar procedure where we train the user-specified ML model using the debiased dataset. However, we use the outcome variable from the original dataset as the ground truth for validation. Sensitive attribute(s) are removed from both datasets before training.
%For the debiased dataset, we train the user-specified ML model using the debiased dataset and then use the original data for validation. More specifically, we use the trained ML model to make predictions for a random subset of the original data, and compute the accuracy and f1 score considering the original output variable as the ground truth. This process is repeated 3 times for different random samples, and the mean accuracy and f1 score is reported. 
Ideally, we would like the utility metrics for the debiased dataset to be close to the corresponding metrics for the original dataset.   
%to make sure that the debiased dataset is approximately as useful as the original dataset. In other words, the accuracy of the classifiers trained over the debiased dataset should be close to the original dataset. We use 3 cross validation to measure the accuracy for the original dataset. The user is allowed to choose a ML model like logistic regression, SVM, etc. We use the debiased dataset to train a machine learning algorithm selected by the user. Thereafter, the trained ML algorithm is used to make predictions on the original dataset. The labels of the original dataset and the model predictions is used to calculate final accuracy. 
%\textcolor{blue}{how debiased accuracy can be higher?}  

\textbf{Data Distortion.} Data distortion is the magnitude of deviation of the debiased dataset from the original dataset. 
%Ideally, we would like to have minimal distortion while optimizing for fairness. 
Since the dataset can have a mix of continuous and categorical variables, we have used the \textit{Gower distance}\cite{gower} metric. 
We compute the distance between corresponding rows of the original and debiased dataset, and then report the mean distance as data distortion. This metric is easy to interpret as it has a fixed lower and upper bound ([0,1]). It is 0 if the debiased dataset is the same as the original while higher values signify higher distortion. Lower values for data distortion are desirable.
%It is represented as a percentage in the donut chart (see Fig \ref{fig:teaser} (C)). 

%which computes separate distance matrices for continuous and categorical variables and then combines them linearly to obtain the final distance matrix. 

%One way can be to first convert all nominal variables to numeric variables by techniques like one hot encoding. Thereafter, we can have use distance metric like euclidean distance, manhattan distance, etc. The downside to this approach is that nominal variables with large number of unique values will dominate other variables. 

\begin{figure}
 \centering 
 \includegraphics[scale=0.25]{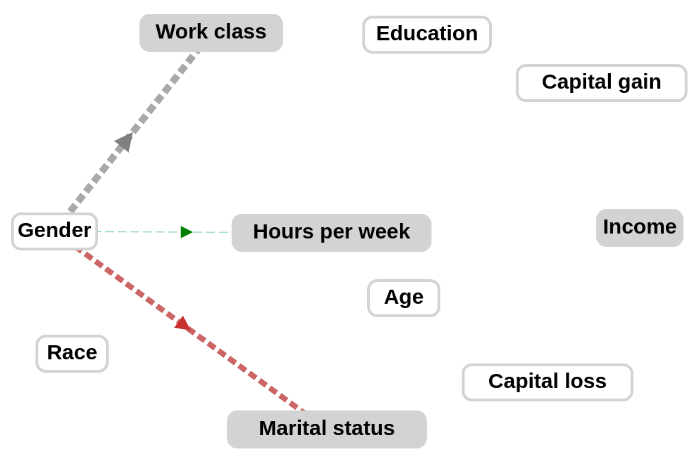}

 \caption{Logs view highlighting the changes made to the causal network for the Adult Income dataset. Dotted lines represent deleted edges and nodes in grey represent the impacted nodes.}
 \label{fig:logs}
\vspace{-10pt}
\end{figure}

%\section{D-BIAS Visual Interface}
\section{The D-BIAS Tool}
%\subsection{Visual Tool}
%In this section, we present the overall system architecture of D-BIAS visual interface by diving into each of its components as shown in Fig.\ref{fig:teaser}.

\subsection{Generator Panel}
This is the first component of the tool the user interacts with (see Fig.\ref{fig:teaser} (A)). The user starts off by choosing a dataset from its dropdown menu. 
%This triggers an event that updates the set of possible options for the label and nominal variables to match the selected dataset. 
The user then selects the label variable which should be a binary categorical variable as we are considering a classification setting. Next, the user selects all nominal variables which is required for fitting the SEM model. Lastly, the user chooses a p-value and clicks the `Causal Model' button to generate the causal network. %(as shown in Fig.\ref{fig:teaser} (B)). 
Here, the p-value is used by the PC algorithm to conduct independence tests. We set $p=0.01$ for all our demonstrations. It can be changed to 0.05 or 0.10 for smaller datasets.
The `Debiased Data' button downloads the debiased dataset. 

\begin{comment}
\begin{figure*}
 \centering 
 \includegraphics[width=2\columnwidth]{figure/teaser_pic.png}
  %\setlength{\belowcaptionskip}{-8pt}
  %\setlength{\abovecaptionskip}{-4pt}
 \caption{The interface of our D-BIAS visual tool using the Adult Income dataset. (A) The Generator panel: used to create the causal network and download the debiased dataset (B) The Causal Network view: shows the causal relations between the attributes of the data, allows user to inject their prior in the system (C) The Evaluation panel: used to choose the sensitive variable, the ML model and displays different evaluation metrics. }  
 \label{fig:teaser}
\end{figure*}
\end{comment}
 
\begin{figure}[t]
 \centering 
 \includegraphics[width=0.75\columnwidth]{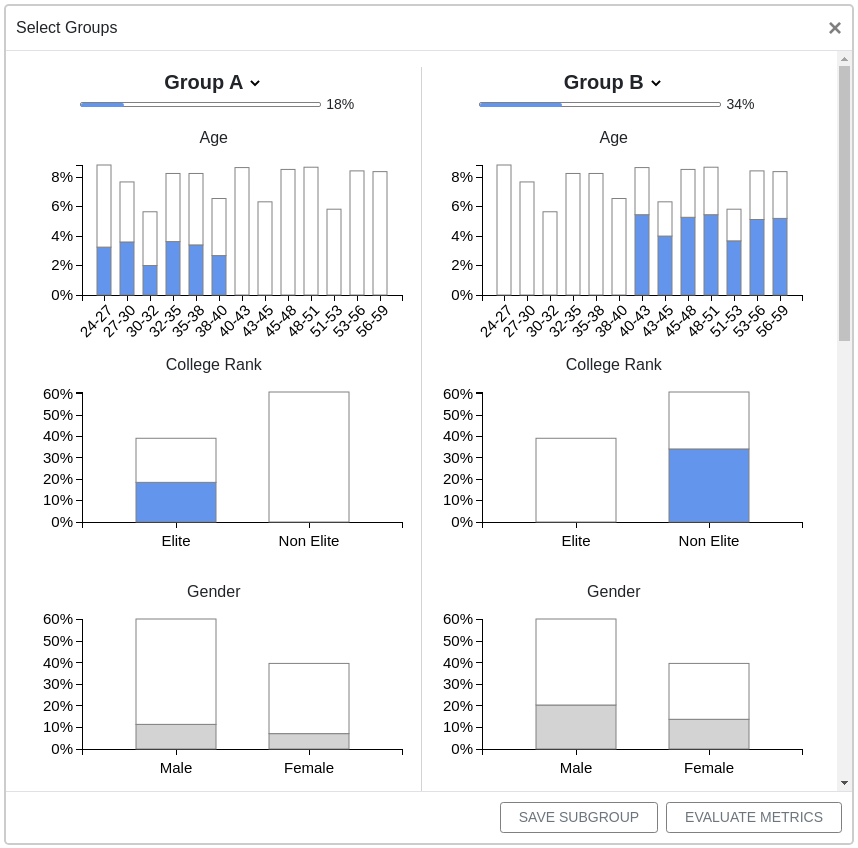}
 \caption{The visual interface for selecting subgroups (Group A and B). Each column consists of a list of bar charts/histograms representing all attributes in the dataset. By default, all bars are colored gray. The user can click on multiple bars to select a subgroup. Selected bars are colored blue. Each bar is filled in proportion of their representation in the selected subgroup as a ratio of the entire dataset. In this picture, Group A consists of individuals who went to elite universities and whose age lies between 24-40. It represents 18\% of the dataset.}  
 \label{fig:custom_group}
\vspace{-10pt}
\end{figure}
 
%After going through the entire workflow, the user will finally return to the generator panel. Here, the user can download the resulting debiased dataset and model by simply clicking on the 'Debiased Data' and 'Download Model' buttons, respectively. 

\subsection{Causal Network View}
This is the most critical piece of the interface where the user will spend most of the time (see Fig.\ref{fig:teaser}(B)). The center of this view contains the actual causal network which is surrounded by 4 panels on all sides. 
%Let's discuss each of these sub-components.\\

\textbf{Causal Network.} All features in the dataset are represented as nodes in the network and each edge represents a causal relation. The width of an edge encodes the magnitude of the corresponding standardized beta coefficient. It signifies the feature importance of the source node in predicting the target node. The color of an edge encodes the sign of the corresponding standardized beta coefficient. Green (red) represents positive (negative) influence of the source node on the target node. If an edge is undirected, it does not have a beta coefficient and is colored orange. Finally, gray color encode edges 
that represent relationships which can not be represented by a single beta coefficient. This occurs when the target node is a categorical variable with more than 2 levels. 

The causal network supports many interactions to enhance the user's overall experience and productivity. It supports operations like zooming and panning. The user can move nodes around if they are not satisfied with the default layout. On clicking a node, all directly connected edges and nodes are highlighted. Similarly, on clicking an edge, its source and target nodes are highlighted. Moreover, clicking a node or an edge also visualizes their distribution in the \textit{Comparison View} (see \autoref{subsec:eval_panel}). 
%If the node represents a numerical feature, its corresponding histogram is displayed in the \textit{Comparison View} (see Fig.\ref{fig:teaser}(E)). Similarly, if the node represents a nominal variable, it's corresponding bar chart is displayed where each bar represents different unique values in that feature. On clicking an edge in the causal network, its source and target nodes are highlighted. Furthermore, the bivariate relationship between the source and target nodes is visualized in the \textit{Comparison View} (see Fig.\ref{fig:teaser}(E)). It can be a scatterplot, grouped bar chart or bar chart with error bars depending on the data types of source and target feature.         

\begin{figure}[h]
\vspace{-1em}
 \centering 
 \includegraphics[width=0.95\columnwidth]{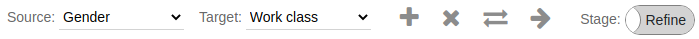}
\vspace{-1 em}
 %\caption{Top chartThis panel allows user to select an edge by choosing source and target nodes. It also allows user to inject bias via } 
 \label{fig:top_chart}
\end{figure}

\textbf{Top panel.} The panel right above the causal network (as shown above) allows the selection of an edge by choosing the source and target nodes. Next to the dropdown menus are a series of buttons which allow a user to inject their prior into the system. Going from left to right, they represent operations like adding an edge, deleting an edge, reversing the edge direction
%revere the sign of the beta coefficient (green to red and vice versa) 
and directing an undirected edge. The toggle at the end represents the current stage (Refine/Debias) and helps transitioning from one to the other.

\begin{figure*}[tb]
 \centering 
 \includegraphics[width=1.85\columnwidth]{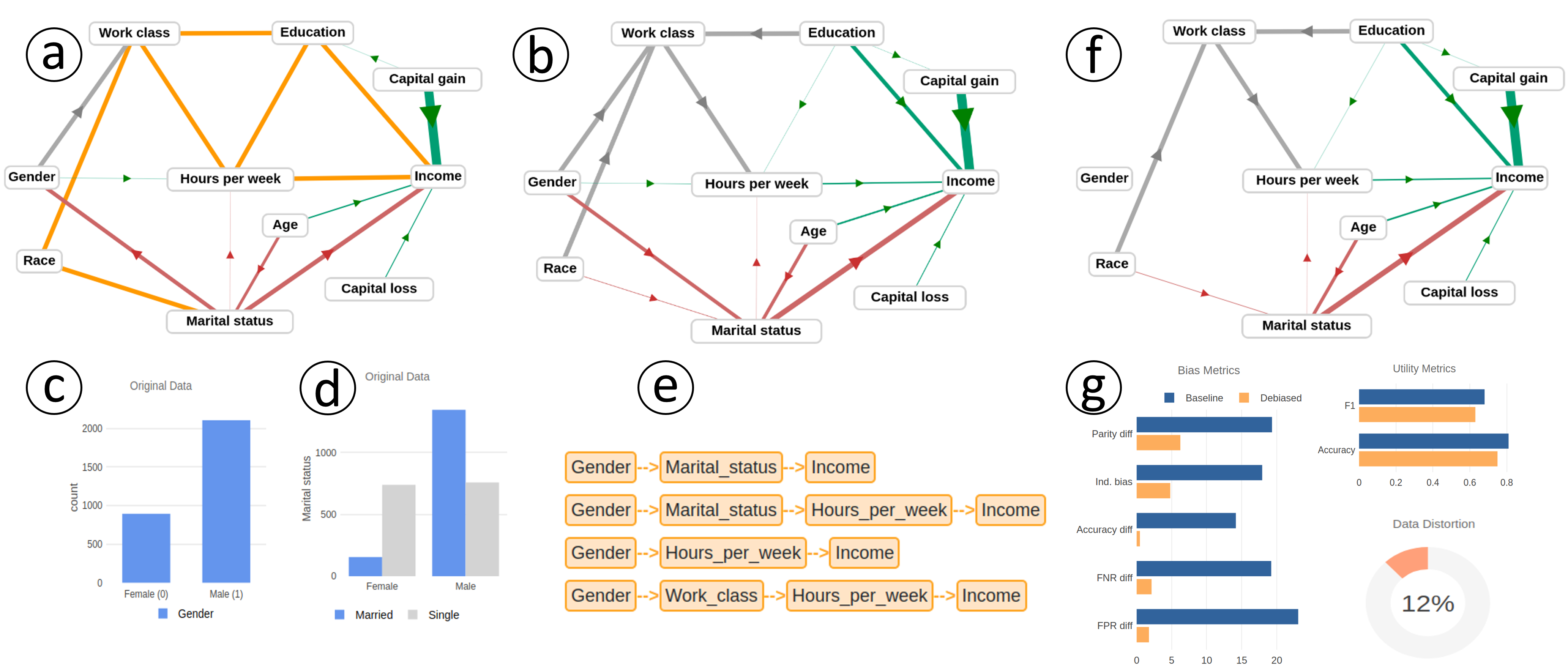}
  \setlength{\belowcaptionskip}{-2pt}
  \setlength{\abovecaptionskip}{-1pt}
 \caption{Case study: Adult Income dataset (a) Causal model generated using automated techniques (b) Refined causal model (c) Clicking the Gender node visualizes its distribution as a bar chart (d) Bivariate distribution between Gender and Marital Status (e) All paths from Gender to Income in the refined causal model (f) Debiased causal model (g) Evaluation metrics to compare our results against the baseline debiasing approach. }  
 \label{fig:adult}
 %\vspace{-10}
\end{figure*}
 
\textbf{Left panel.} The bar to the left of the causal network shows the change in BIC score. This bar is updated each time the user performs operations like directing an undirected edge, adding/deleting an edge, etc. A negative value means that the change made to the causal network is in sync with the underlying dataset; positive values mean the opposite. Negative (positive) values are represented in green (red).   

\textbf{Right panel.} The panel to the right of the causal network offers four functionalities. Going from top to bottom, they represent zooming in, zooming out, reset layout and changing weight of an edge. 
%D-BIAS supports zooming in and out via buttons apart from mouse scrolling to support uniform zooming across different hardware. 
The slider at the bottom gets activated when the user clicks on an edge during the debiasing stage. It allows the user to weaken/strengthen an edge depending on the selected value between -100\% to 100\%. Moving the slider changes $\alpha_i$ and also impacts the effective beta coefficient for the selected edge ($\alpha_i\beta_i$). This change manifests visually in the form of proportional change in the corresponding edge width. Moving the slider to -100\% will result in deletion of the selected edge.

%\vspace{-1 em}
\begin{figure}[h]
 \centering 
 \includegraphics[scale=0.2]{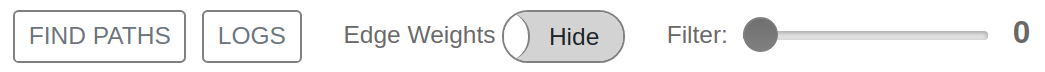}
\vspace{-2 em}
 %\caption{} 
 \label{fig:bottom_chart}
\end{figure}

\textbf{Bottom panel.} As shown below, the bottom panel offers 4 functionalities. The ``Find Paths" button triggers depth first traversal of the causal graph to compute all directed paths between the source and target node as selected in the top panel. This will be especially helpful %to find all paths between the sensitive and the label variable 
when the graph is big and complex. All the computed paths are then displayed below the bottom panel. A user can click on a displayed path to highlight it and see an animated view of the path going from the source to the target node. The ``Logs" button highlights changes made to the causal network during the debiasing stage (see Fig.\ref{fig:logs}). All edges are hidden except for the newly added edges, deleted edges and edges that were weakened/strengthened. Nodes impacted by such operations ($V_{sim}$) are highlighted in grey. 
%In Fig.\ref{fig:logs}, we see how the causal network looks like after "Logs" button is clicked. 
%All edges are hidden and only modified edges are shown. 
%Here, two edges are highlighted because those two were deleted by the user. Deleted edges are represented by dotted lines. 
%This view also represents all nodes impacted by the changes made to the edges. Background color for \textit{Major} and \textit{Job} represents that these nodes will be modified to account for the two deleted edges. Nodes with white background color will remain untouched.

If a user is interested in knowing the exact beta coefficients for all edges, the edge weights toggle will help in doing just that. By default, it is set to `hide' to enhance readability. If turned to `show', the beta coefficients will be displayed on each edge. The filter slider helps user focus on important edges by hiding edges with absolute beta coefficients less than the chosen value. 
%This feature plays a key role in dealing with large causal networks.   

\subsection{Evaluation Panel}
\label{subsec:eval_panel}
This panel, at the right (see Fig.\ref{fig:teaser}(C)), provides different options and visual plots for comparing and evaluating the changes made to the original dataset. From the top left, users can select the sensitive variable and the ML algorithm from their respective dropdown menus. This selection will be used for computing different fairness and utility metrics.  
%based on which fairness metrics will be computed. Users can also specify the ML algorithm used to compute the classification based fairness metrics and utility metrics like accuracy.
For the sensitive variable, the dropdown menu consists of all binary categorical variables in the dataset along with a \textit{Custom Group} option. Selecting the \textit{Custom Group} option opens a new window (see Fig. \ref{fig:custom_group}) which allows the user to select groups composed of multiple attributes. This interface facilitates comparison between intersectional groups say black females and white males. 
%To see the impact of the changes made to the causal network in terms of different evaluations metrics, the user can simply click on the "Evaluate Metrics" button.

Clicking the ``Evaluate Metrics" button triggers the computation of evaluation metrics that are displayed on the right half of this panel (Performance View). It  also visualizes the relationship between the sensitive attribute or selected groups and the outcome variable using a 4-fold display \cite{friendly1994fourfold} on the left half of this panel (Comparison View). 
%(see \autoref{fig:teaser}. 

\textbf{Comparison View.} This view comprises two plots aligned vertically where the top plot represents the original dataset and the bottom plot represents the debiased dataset. 
%this view also helps compare the changes made to individual data attributes and binary relationships between them. 
This view has two functions. It first aids the user in the initial exploration of different features and relationships. When a node or edge is clicked in the causal network, the summary statistics of the corresponding attributes is visualized (see Fig. \ref{fig:adult}(c) for an example). For binary relationships, we use either a scatter plot, a grouped bar chart or an error bar chart depending on the data type of the attributes. The second function of the Comparison View is to visualize the differences between the original and the debiased dataset. Initially, the original data is the same as the debiased data and so identical plots are displayed. However, when the user injects their prior into the system, the plots for the original and debiased datasets start to differ. We added this view to provide more transparency and interpretability to the debiasing process and also help detect sampling bias in the data. 

%Please refer to the supplemental material to see these visualizations. 
When the user clicks the ``Evaluate Metrics" button, the Comparison View visualizes the binary relation between the sensitive attribute or selected groups and the outcome (label) variable via the 4-fold plot~\cite{friendly1994fourfold} as shown in Fig.\ref{fig:teaser}(C), left panel). 
%The second function of the Comparison View is to visualize the differences between the original and the debiased dataset generated in the simulation via the 4-fold plot as shown in Fig. \ref{fig:teaser}(C), left panel).  
We chose a 4-fold display over a more standard  brightness-coded confusion matrix since the spatial encoding aids in visual quantitative assessments. The left/right half of this display represents two groups based on the chosen sensitive variable (say males and females) or as defined in the Custom Group interface, while the top/bottom half represents different values of the output variable say getting accepted/rejected for a job. Here, symmetry along the y-axis means better group fairness.

%On evaluation,  this view is used to display the original and debiased dataset as a 2D scatterplot where each point is color-coded with the sensitive variable. 

\textbf{Performance View.}
This view houses all the evaluation metrics as specified in Sec.\ref{sec:eval_metrics}. It uses a horizontal grouped bar chart to visualize 2 utility and 5 fairness metrics. Lower values for the fairness metrics mean better fairness. Higher values for utility metrics means better utility. Data distortion is visualized using a donut chart. On hovering over any of these charts, a tooltip shows the exact values.   

\section{Case Study}
\label{sec:case_study}
We demonstrate the utility of our tool for bias identification and mitigation using the Adult Income dataset. Each data point in the dataset represents a person described by 14 attributes recorded from the US 1994 census. Here, the prediction task to classify if a person's income will be greater or less than \$50k based on attributes like age, sex, education, marital status, etc. We chose this dataset as it is widely used in the algorithmic fairness literature \cite{inprocess,nipsIBM,ghai2022cascaded}. Here, we have chosen a random sample of 3000 points from this dataset for faster computation. 
%In this use case, we will try to mitigate the influence of gender on income with the strategies discussed in Section \ref{debiaisngTech}

\begin{figure*}
 \centering 
 \includegraphics[width=1.9\columnwidth]{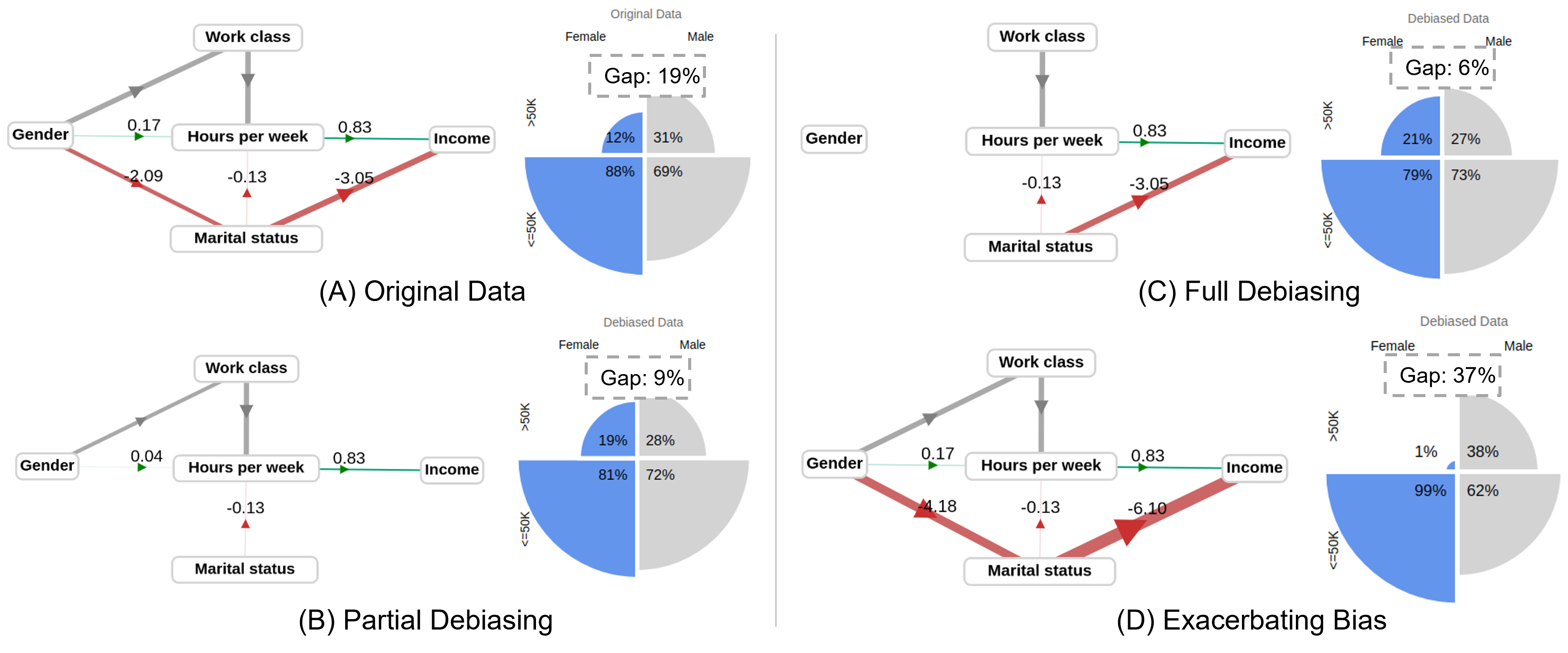}
  \setlength{\belowcaptionskip}{-8pt}
  \setlength{\abovecaptionskip}{-4pt}

 \caption{The above picture shows the impact of 3 types of user interaction as captured by the 4-fold display. Due to space constraints, we have only shown a subset of the causal network which connects the Gender node with the Income node. For details, please refer to the description in Section \ref{sec:case_study}.} 
 \label{fig:adult_fourfold}
\end{figure*}

\textbf{Generating the causal network.}
We start off by selecting the Adult Income dataset from the respective dropdown menu in the Generator panel. 
%On selecting the dataset, the options for the label variable and nominal variable dropdown are updated.
We select \textit{Income} as the label variable and \textit{Work class}, \textit{Marital Status}, \textit{Race}, \textit{Gender}, \textit{Income} as the nominal variables. Next, we click on the \textit{Causal Model} button which generates the default causal model (see Fig. \ref{fig:adult} (a)). 
%As we are in the Refine stage, 
Here, we examine different edges of the causal model and act on them as needed to reach to a reliable causal model. We start with the 7 undirected edges encoded in orange. We direct edges based on our domain knowledge like \textit{Hours per Week} $\to$ \textit{Income},
%\textit{Race} $\to$ \textit{Work class}, 
\textit{Education} $\to$ \textit{Income}, \textit{Education} $\to$ \textit{Hours per week}, etc. After each of these operations, we observe a green bar in the left panel of the Causal Network View. This indicates that the resulting causal model is a better fit over the underlying dataset.  
%For example, we observe undirected edges like \textit{income} $\leftrightarrow$ \textit{Hours per Week}, \textit{Race} $\leftrightarrow$ \textit{Work class}, \textit{Education} $\leftrightarrow$ \textit{Income}, \textit{Hours per week} $\leftrightarrow$ \textit{Education}, etc. 
%Here, we can choose to direct these edges or delete them. In this case, we choose to direct all these edges
Next, we examine other directed edges. Many of them align with our domain knowledge like \textit{Capital Gain} $\to$ \textit{Income}, \textit{Age} $\to$ \textit{Income}, etc. However, we found a couple of them to be counter-intuitive, namely \textit{Capital Gain} $\to$ \textit{Education} and \textit{Marital Status} $\to$ \textit{Gender}. In a causal relation, cause always precedes effect. Hence, immutable personal characteristics like sex, race, etc., which are assigned at birth, can not be the effect of a later life event like marriage or work class. So, in this case, we chose to reverse both these edges to get to the refined causal model (see Fig. \ref{fig:adult}(b)). 
%These edits will not cause any change to the dataset since we are in the Refine stage.         
%After loading the dataset, \textit{income} is selected as the label attribute and \textit{sex} as the protected variable. \textit{Race}, \textit{relationship}, \textit{marital status} and \textit{work class} are the nominal variables in the dataset. We built the influence network shown in \autoref{fig:adultMDS}(a)), which describes the underlying data dependencies and how the bias is propagated. 

\textbf{Auditing for social biases.}
Once we have reached a reliable causal model, we start auditing for different kinds of biases. We click on different nodes and edges to explore their distributions. For eg., clicking the \textit{Gender} node visualizes its distribution in the \textit{Comparison View}. We observe that females are underrepresented in the dataset (895 females vs 2105 males) (see Fig. \ref{fig:adult} (c)). 
%Such representation bias can potentially lead to algorithmic bias 
Given this representation bias and the fact that gender pay gap is a well-known issue, we decided to investigate further.  
We found an indirect path from \textit{Gender} to \textit{Income} via \textit{Hours per week}. This indicates a possible disparity in income based on gender. To probe further, we click the ``Evaluate Metrics" button with \textit{Gender} as the sensitive variable to compute different fairness metrics. We observe significant gender bias as captured by metrics like Accuracy diff (14\%), FPR diff (22\%), FNR diff (17\%), etc.
%The Baseline visualization in Fig.~\ref{fig:adult}(f) (blue bars) reports a large disparity in Accuracy diff (10.8\%), FPR diff (9\%), FNR diff (24.4\%), etc. 
The 4-fold display for the original data in Fig. \ref{fig:adult_fourfold} (A) reveals that only 12\% of the females earn more than \$50k compared to 31\% for males. Thus there is a significant income disparity based on gender. To have a more comprehensive understanding of the issue, we search for all possible paths by selecting \textit{Gender} and \textit{Income} as the source and target from the Top panel and then clicking the ``Find Paths" button from the bottom panel. This populates 4 different causal paths below the bottom panel (see Fig. \ref{fig:adult} (e)). We will now focus on the different causal relationships in these paths and try to make minimal changes to achieve more fairness.      

\textbf{Bias mitigation.} To mitigate gender bias, we first enter the debiasing stage by flipping the Stage toggle from Refine to Debias. From here on, any changes to the causal network will simulate a new debiased dataset. 
%with its impact on fairness visualized in the Debiased Data 4-fold display. 
Among the 4 paths we previously discovered, the top 2 paths have a common causal edge, i.e., \textit{Gender} $\to$ \textit{Marital status}. On clicking this edge, we find that most males in the dataset are married while most females are single (see Fig. \ref{fig:adult}(d)). This pattern indicates sampling bias.
%as it doesn't existing in the real world. 
Ideally, we would like no relation between these attributes so we delete this causal edge. 
%This operation will trigger a change for \textit{Marital status} and its children nodes i.e., \textit{Hours per week} and \textit{Income}. We can assess the overall impact of this operation by clicking the "Evaluate Metrics" button. We observe no change in the utility metrics, but a significant reduction across all fairness metrics and a 7\% data distortion. 
%Furthermore, the percentage of females earning $>$\$50k increases from 12\% to 17\% while for males it comes down from 31\% to 29\% (see Fig. \ref{fig:adult_fourfold}). Hence, decreasing the gender gap. 
%If one is satisfied with these results, they can just stop here and download the debiased dataset. Otherwise, they can continue the debiasing process for potentially achieving more fairness at the cost of higher data distortion.
%and deal with remaining unfair causal paths. 
%move on on and deal with the remaining 2 causal paths for 
Next, we assess the remaining two causal paths. Based on our domain knowledge, we find the causal edge \textit{Hours per week} $\to$ \textit{Income} to be socially desirable, and the edges \textit{Gender} $\to$ \textit{Work class} and \textit{Gender} $\to$ \textit{Hours per week} to be socially undesirable. We delete the two biased edges to get to the debiased causal model (see Fig. \ref{fig:adult}(f)). To verify all changes made so far, we click on the \textit{Logs} button. As shown in Fig. \ref{fig:logs}, it shows 3 dotted lines for the removed edges and highlight the impacted attributes. 

Lastly, we click on the ``Evaluate Metrics" to see the effect of our interventions. The 4-fold display for the debiased data in Fig. \ref{fig:adult_fourfold}(C) shows that the disparity between the two genders has now decreased from 19\% to 6\% (compare Fig. \ref{fig:adult_fourfold}(A)). The percentage of females who make more than \$50k have undergone a massive growth of 75\% (from 12\% to 21\%). 
%This amounts to a massive growth of 75\% for females who make more than \$50k group, at a relatively small drop of 13\% for the male group. 
%Clicking the ``Evaluate Metrics" button generates the Performance View in Fig.~\ref{fig:adult}(f). Here, Baseline (blue bars) are the metrics produced by the ML model trained with the Original data but without the sensitive \textit{Gender} attribute, which is the conventional (baseline) debiasing strategy. We observe a fairly large disparity in Accuracy diff (10.8\%), FPR diff (9\%), FNR diff (24.4\%), etc. In contrast, Debiased (orange bars) are the metrics produced by the ML model trained with the debiased data with proxy bias removed (see Section 3.3). 
Moreover, as shown in Fig.~\ref{fig:adult}(g), we find that all fairness metrics have vastly improved, with only a slight decrease in the utility metrics and an elevated distortion (12\%). 
%Along with the 4-fold plot analysis described above, this metric-based evaluation clearly indicates that our method produces superior debiasing results.
These results clearly indicate the efficacy of our debiasing approach.

%but all fairness metrics have vastly improved (see the Debiased visualization in Fig. \ref{fig:adult} (f) (orange bars) and \autoref{tab:table}). 

%whereby the Males experience only  a modest drop of 13\%. 
%Finally, the debiased dataset can be downloaded by clicking the 'Debiased Data' button from the Generator panel.  

\textbf{Partial debiasing.} Considering the tradeoff between different metrics, one might choose to debias data partially based on their context, i.e., weaken biased edges instead of deleting them or keeping certain unfair causal paths from the sensitive variable to the label variable intact.
For eg., one might choose to delete the edge \textit{Gender} $\to$ \textit{Marital status} and weaken the edges \textit{Gender} $\to$ \textit{Work class} and \textit{Gender} $\to$ \textit{Hours per week} by 25\% and 75\%, respectively. On evaluation, we find this setup to sit somewhere between the original and the fully debiased case (see Fig. \ref{fig:adult_fourfold} (B)). It performs better on fairness than the original dataset (gap: 9\% vs 19\%) but worse than the full debiased version (gap: 9\% vs 6\%). Similarly, it incurs more distortion than the original dataset but less than the full debiased version (11\% vs 12\%).

\begin{table*}
  \caption{Evaluation metrics to compare the debiased dataset generated using our tool against the baseline debiasing approach for different datasets.}
  \label{tab:table}
  \scriptsize%
	\centering%
  \begin{tabu}{%
	r%
	*{10}{c}%
	*{2}{c}%
	}
  \toprule
   \multirow{2}{*}{Dataset} & Sensitive & \multirow{2}{*}{version} & \multirow{2}{*}{ML model} & \multirow{2}{*}{Accuracy} & \multirow{2}{*}{F1} & Parity & Individual & Accuracy &  FNR &  FPR & Data \\ 
   & attribute & & & & & difference & Bias & difference &  difference &  difference & Distortion \\ 
  \midrule \\
  \multirow{2}{*}{Synthetic Hiring} & \multirow{2}{*}{Gender} & baseline & \multirow{2}{*}{SVM} & 77\% & 0.59 & 11.12 & 19.09 & 4.14 & 14.26 & 6.82 & 0\% \\ 
  & & debiased & & 77\% & 0.60 & 1.66 & 12.93 & 2.99 & 1.37 & 3.63 & 6\% \\ \\
  
  \multirow{2}{*}{Adult Income} & \multirow{2}{*}{Gender} & baseline & Logistic & 82\% & 0.69 & 19.32 & 17.92 & 14.35 & 17.98 & 22.53 & 0\%\\ 
  & & debiased & Regression & 75\% & 0.63 & 6.24 & 4.8 & 0.88 & 2.33 & 1.9 & 12\% \\ \\
  
  \multirow{2}{*}{COMPAS} & \multirow{2}{*}{Race} & baseline & Random & 67\% & 0.64 & 12.07 & 33.9 & 0.17 & 23.07 & 16.89 & 0\%\\
  & & debiased & Forest & 63\% & 0.59 & 11.12 & 2.19 & 0.44 & 0.68 & 1.55 & 13\% \\ \\
  
  \bottomrule
  \end{tabu}%
\vspace{-5pt}
\end{table*}

\textbf{Intersectional groups.} D-BIAS facilitates auditing for biases against intersectional groups using the ``Custom Group" option from the sensitive variable dropdown.
% (see Fig. \ref{fig:custom_group})
%Clicking this option opens a new window which allows selection of subgroups based on different constituting attributes (see Fig. \ref{fig:custom_group}). 
%In the case of Adult Income dataset, one might 
Here, we choose \textit{Black Females} and \textit{White Males} as the two groups. At the outset, there is a great disparity between the groups as reflected in the 4-fold display and the fairness metrics (see Fig. \ref{fig:teaser}). 
As these subgroups are defined by \textit{Gender} and \textit{Race}, we focus on the unfair causal paths from these nodes to the label variable (\textit{Income}). 
%For full debiasing, we need to break all such causal paths.
%and arrive at the causal model shown in Fig. \ref{fig:teaser}. 
For debiasing, we first perform the same operations we did for gender debiasing. Thereafter, we reduce the impact of race by deleting the edges \textit{Race} $\to$ \textit{Work class} and \textit{Race} $\to$ \textit{Marital status} which we deem as socially undesirable. On evaluation (see Fig. \ref{fig:teaser}), we find a significant decrease in bias across all fairness metrics for the debiased dataset compared to the conventional debiasing practice (blue bars) which just trains the ML model with the sensitive attributes (here \textit{Gender} and \textit{Race}) simply removed.
%slight decrease in utility metrics and a 12\% data distortion. 
Finally, the two 4-fold displays reveal that the participation of the disadvantaged group more than doubled, while the privileged group experienced only a modest loss.

\textbf{Exacerbating bias.} The flexibility offered by D-BIAS to refine the causal model can be misused to increase bias as well. 
%There can be applications where this seems reasonable but these typically do not operate on sensitive variables. 
Bias can be exacerbated by strengthening/adding biased causal edges and weakening/deleting other relevant causal edges.
For eg., 
%in the case of Adult Income dataset, 
one can exacerbate gender bias by strengthening the edges \textit{Gender} $\to$ \textit{Marital status} and \textit{Marital status} $\to$ \textit{Income} by a 100\%. On evaluation, we find that the proportion of females making $>$\$50k has shrunk to just 1\% while the proportion of males has surged to 38\%. In effect, this has broadened the gap between males and females making more than \$50k by about 2x from 19\% to 37\% (see Fig. \ref{fig:adult_fourfold} (D)). 
%Here, we have strengthened causal edges belonging to one causal path from the sensitive attribute to the outcome variable. Similarly, one can strengthen causal edges belonging to other causal paths to increase bias even further.  

\textbf{Results.} Apart from the Adult Income dataset, we also tested our tool using the synthetic hiring dataset and the COMPAS recidivism dataset (see appendix D for details). 
%We also tested our tool on the Adult Income dataset\cite{aif360} for gender bias and the German Credit dataset\cite{aif360} for age bias. 
%Our tool performed well in identifying and mitigating bias for both of these datasets. 
The evaluation metrics for all 3 datasets after full debiasing are reported in Table \ref{tab:table}. As we can observe, our tool is able to reduce bias significantly compared to the baseline approach across the 3 datasets for a small loss in utility and data distortion. \textit{These results validate the potential of HITL approach in mitigating bias}. It is interesting to observe that the F1 score for the synthetic hiring dataset is slightly higher than the baseline approach. However, this is line with the existing literature \cite{ghai2022cascaded} where similar instances have been recorded.

\section{User Study}
\label{sec:user_study}
We conducted a user study to evaluate two primary goals: (1) usability of our tool, i.e., if participants can comprehend and interact with our tool effectively to identify and mitigate bias, (2) compare our tool with the state of the art in terms of human-centric metrics like accountability, interpretability, trust, usability, etc.
%We conducted a user study to evaluate the usability of our tool i.e., for first identifying and then mitigating bias. We aimed to see if participants can comprehend and interact with our interface effectively to yield desired results. We were also interested in how our tool compares with automated ML algorithms in terms of usability, interpretability, trust.

\textbf{Participants.} We recruited 10 participants aged 24-36; gender: 7 Male and 3 Female; profession: 8 graduate students and 2 software engineers.
%8 are graduate students and 2 of them work as software engineers for reputed firms. 
The majority of the participants are computer science majors with no background in data visualization or algorithmic fairness. 80\% of the participants trust AI and ML technologies in general.
The participation was voluntary with no compensation. 

\begin{figure}[t]
 %\vspace{-10pt}
 \centering 
 \includegraphics[scale=0.50]{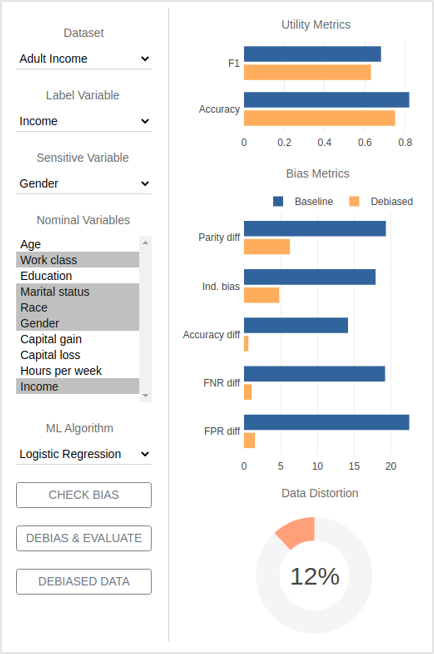}
 \vspace{-8pt}
 \caption{Baseline visual tool used as a benchmark in the user study.}
 %which mimics the workflow of IBM's AI Fairness 360 toolkit and possesses the same visual design as the D-BIAS tool.
 \label{fig:baseline}
 %\vspace{-10}
\end{figure}

\textbf{Baseline Tool.} For an even comparison, we looked for existing tools with a visual interface that support bias identification and mitigation. This led us to IBM’s AI Fairness 360 \cite{aif360} toolkit whose visual interface
%which contains state of the art ML algorithms to examine and mitigate bias across different stages of the AI pipeline. It also has a visual interface 
can be publicly accessed online\footnote{https://aif360.mybluemix.net/data}. However, we didn't go further with this toolkit as the baseline (control group) because it has a significantly different look and feel which is difficult to control for. Instead, we took inspiration from this toolkit and built a baseline visual tool (not to be confused with the baseline debiasing strategy) which mimics its workflow but matches the design of our D-BIAS tool (see Fig.\ref{fig:baseline}).

IBM’s AI Fairness toolkit allows the user to choose from a set of fairness enhancing interventions with varying impact on the evaluation metrics. However, this study is focused on other important aspects such as trust, accountability, etc. So, in order to have a tightly controlled experiment, we imagine a hypothetical automated debiasing algorithm whose performance exactly matches the peak performance of our tool for all evaluation metrics. Here, peak performance refers to the state where all unfair causal edges are deleted. %Furthermore, we made sure that the baseline tool supports the exact same datasets and evaluation metrics as the D-BIAS tool. 

Using the baseline tool is quite simple (see appendix E). The user first selects the dataset, label variable, etc. 
%in the same fashion as the D-BIAS tool. 
They can then audit for different biases by selecting the sensitive attribute and then clicking on the `Check bias' button. This will compute and present a set of fairness metrics in the same fashion as the D-BIAS tool. Lastly, the user can click the `Debias \& Evaluate' button to debias the dataset and generate its evaluation metrics. A small lag is introduced before displaying evaluation metrics to mimic a real debiasing algorithm. 

%There are different automated debiasing algorithms supported by IBM's AIF 360 whose performance varies over different fairness metrics. Choosing one of them for the baseline tool will lead to differences in evaluation metrics from our own tool. This difference will in turn impact the user experience and will be harder to control for. 

\textbf{Study design.}  We conducted a within subject study where each participant was asked to use the baseline tool and D-BIAS in random order. The study was conducted remotely, i.e., each participant could access and interact with the tools via their own machine. For each tool, a small tutorial was given using the Synthetic Hiring dataset\footnote{We generated a synthetic hiring dataset fraught with gender and racial bias to better evaluate our tool.
%as the underlying data generation process is known. 
For details, please refer to appendix C .} to demonstrate the workflow and features of the tool. Each participant was then given some time to explore and interact with each system. Next, the participants were asked to identify and mitigate bias for the Adult Income dataset. 
%We observed where the participants struggled or failed to perform a task and why it happened.  
%Their verbal accounts can help gather actionable intelligence to further refine the tool. 
%The study was organized in three phases. The first phase is the tutorial phase where each participant was briefed about the overall theme and objective of our tool. 
For the D-BIAS tool, participants were also asked to complete a set of 5 tasks to evaluate usability. Each task was carefully designed to cover our testing goals and had a verifiable correct solution.
%This is required to check if a participant has successfully completed that task. 
Tasks included: generate a causal network, direct undirected edges, identify if bias exists with respect to an attribute, identify proxy variables and finally debias the dataset. After using each tool, the participants were asked to answer a set of survey questions. Lastly, we collected subjective feedback from each participant regarding their overall experience with both the tools. 
Throughout the study, participants were in constant touch with the moderator for any assistance.
Participants were encouraged to think aloud during the user study. 
%The moderator could listen and observe all interactions made by each participant. 

\textbf{Survey Design.} Each participant was asked to answer a set of 13 survey questions to quantitatively measure usability, interpretability, workload, accountability and trust. All of these questions can be answered on a 5-point Likert Scale. To capture cognitive workload, we selected two applicable questions from the NASA-LTX task load index \cite{nasa_tlx}, i.e., ``How mentally demanding was the task? and ``How hard did you have to work to accomplish your level of performance?". Participants could choose between 1 = very low to 5 = very high. For capturing usability, we picked 3 questions from the System Usability Scale (SUS) \cite{sus}. For e.g., ``I thought the system was easy to use", ``I think that I would need the support of a technical person to be able to use this system". Participants could choose between 1 = Strongly disagree to 5 = Strongly agree. To capture accountability, we asked two questions based on previous studies \cite{XAL, cai2019effects}. For e.g., ``The credit/blame for mitigating bias effectively is totally due to" (1 = System's capability, 5 = My input to the system). To capture interpretability, we consulted Madsen Gregor scale\cite{madsen2000measuring} and adopted 3 questions for our application. For e.g., ``I am satisfied with the insights and results obtained from the tool?", ``I understand how the data was debiased?" Answers could lie between 1 = Strongly disagree to 5 = Strongly agree. For measuring trust, we referred to McKnight's framework on Trust\cite{trust1, trust2} and other studies \cite{XAL, drozdal2020trust} to come up with 3 questions for our specific case. For e.g., ``I will be able to rely on this system for identifying and debiasing data" (1 = Strongly disagree, 5 = Strongly agree). 
%To have an even comparison we looked for other tools which helps in bias identification, mitigation and also possess a visual interface. We compare D-BIAS with IBM AIF360 tool. \\
% 

\textbf{Results.} Despite not having a background in algorithmic fairness or data visualization, all participants were able to complete all 5 tasks using the D-BIAS tool. This indicates that our tool is easy to use. 
%given a tutorial in the beginning.
%We observed that all participantsre able to reduce bias by different extents. But there was variab \\

\begin{figure} [t]
% \centering 
  %\vspace{-15pt}
 \includegraphics[width=\columnwidth]{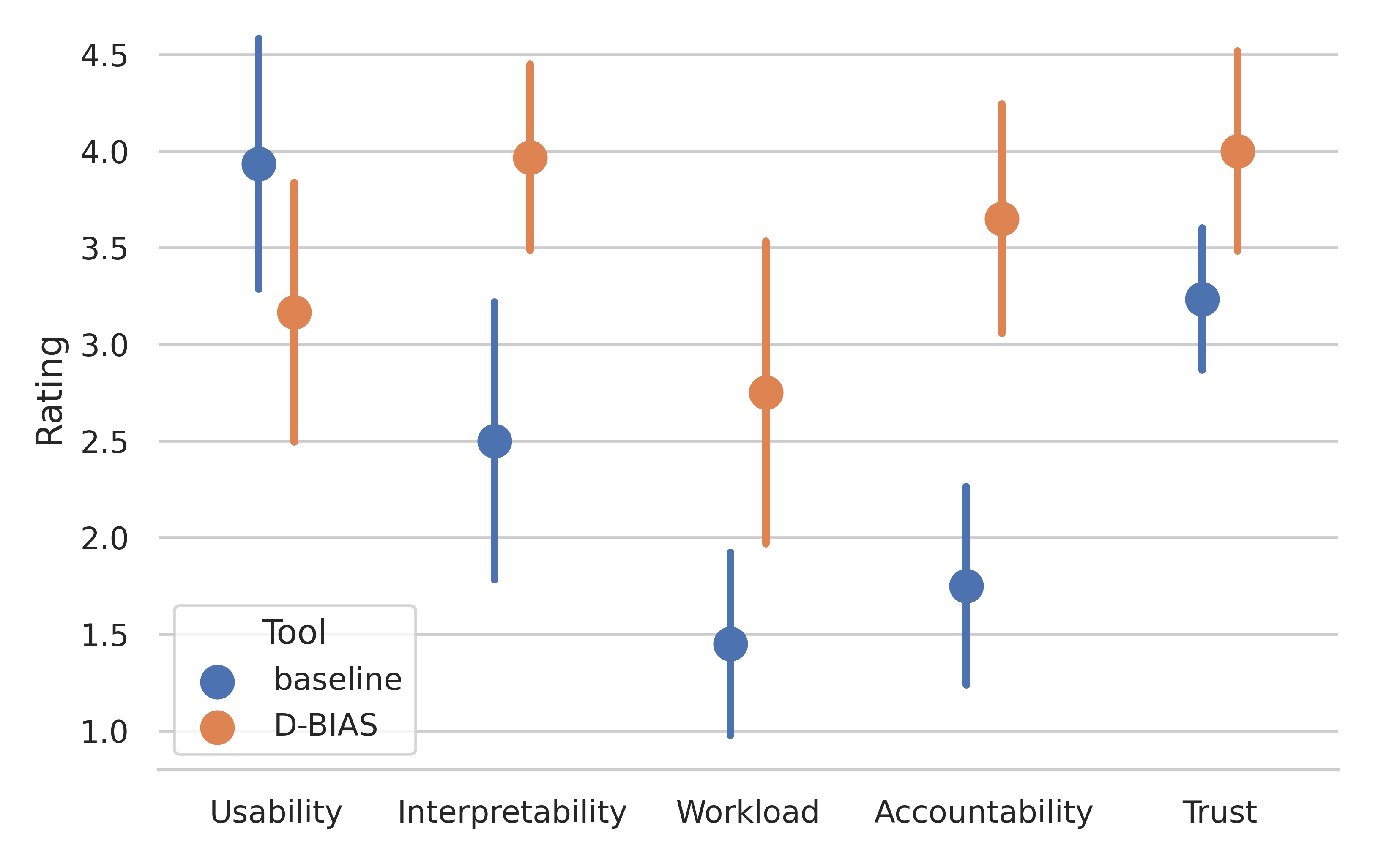}
  \setlength{\belowcaptionskip}{-8pt}
\setlength{\abovecaptionskip}{-4pt}
 \vspace{-1em}
 \caption{Mean user ratings from the survey data along with their standard deviation for different measures. %Differences in mean ratings are statistically significant with $p<0.05$
 } 
 \label{fig:user_study}
\end{figure}

  The survey data was analyzed to calculate usability, interpretability, workload, accountability and trust ratings for each tool by each participant. The mean ratings along with their standard deviations are plotted in Fig.\ref{fig:user_study}. 
  %We used t-test to check if the differences in ratings are significant. 
  Using t-test, we found statistically significant differences for all measures with p$<$0.05. \textit{We found that D-BIAS outperforms the baseline tool in terms of trust, accountability and interpretability.} However, it lags in usability and cognitive workload.
  %The baseline tool performs better in usability and cognitive workload because there is not a lot that the user can do. 
  So, if someone is looking for a quick fix or relies on ML algorithms more than humans, automated debiasing is the way to go. Conversely, if trust, accountability or interpretability is important, D-BIAS should be the preferred option. Looking at these results in conjunction with the results reported in Table 1, \textit{we find that our tool enhances fairness while fostering accountability, trust and interpretability.}     
  %Our approach might require the user to go through a tutorial and spend some time So, if Trust, Interpretability or Accountability  This is a strong validation for our human in the loop methodology vs automated debiasing approach.   

\textbf{Subjective feedback.}
After the study, we gathered feedback from each participant about what they liked or disliked about D-BIAS. Most participants liked the overall design, especially the causal network. We got comments like ``\textit{Interface is user friendly}", ``\textit{Causal network gives control and flexibility}", ``\textit{Causal network is very intuitive and easy to understand}", ``\textit{Causal network is a great way to understand relationships between features}". Most participants agreed that after a tutorial session, it should be fairly easy for even non-experts to play with the system. Another important aspect which received a lot of attention was our human-in-the-loop approach. Participants felt that they had a lot more control over the system and that they could change things around. One of the participants commented ``\textit{It feels like I have a say}". Some of the participants said they felt more accountable because the system offered much flexibility and that they had a choice to make.

Many of the participants strongly advocated for D-BIAS over the baseline tool. For e.g., ``\textit{D-BIAS better than automated debiasing any day}", ``\textit{D-BIAS hand's down!}". Few of the participants had a more nuanced view. They were of the opinion that the baseline tool might be the preferred option if someone is looking for a quick fix. 
We also received concerns and suggestions for future improvement. Two of the participants raised concern about the tool's possible misuse if the user is biased. Another participant raised concern about scalability for larger datasets. Most participants felt that adding tooltips for different UI components especially the fairness metrics will be a great addon. Two participants wished they could see the exact changes in CSV file in the visual interface itself. 
%Considering survey data results along with subjective feedback provides a strong validation for human in the loop methodology for tackling algorithmic bias.

%\section{Discussion}
\section{Discussion, Limitations, and Future work}
\textbf{Efficacy.} The efficacy of our tool depends on how accurately the causal model captures the underlying data generating process and the ensuing refining/debiasing process. As our tool is based on causal discovery algorithms, it inherits all its assumptions such as the causal markov condition and its limitations like sampling biases, etc.\cite{challenges}.
%Every variable X in V (the set of variables in the causal graph)is independent of its non-effects conditional on its direct causes, etc.
%The PC algorithm assumes that all confounder variables are part of the dataset. This is not always the case and hence can lead to erroneous causal edges. \textcolor{blue}{We observed this in the case of the Synthetic Hiring dataset where the causal discovery algorithm} falsely detected an edge between \textit{SAT score} and \textit{Grade point Average}. 
%Similarly, it is also susceptible to sampling biases. 
For eg., Caucasians had a higher mean age than African Americans in the COMPAS dataset. So, the PC algorithm falsely detected a causal edge between \textit{Age} and \textit{Race} (see appendix D.2). From our domain knowledge, we can deduce that this error is due to sampling bias. %and not directly linked with historical biases. 
Ideally, one should use such an insight to gather additional data points for the under sampled group. However, it may not always be possible. In such cases, our tool can be leveraged to remove such patterns in the debiasing stage. 
We dealt with a similar case (\textit{Gender} $\to$ \textit{Marital status}) for the Adult Income data (see Sec. \ref{sec:case_study}). 
%In the interest of minimizing data distortion, we recommend this operation only when the specific causal edge lies on an unfair causal path between the sensitive and the outcome variable.  
%Such errors can be easily dealt with in the refining stage. 
%However, there can be other subtle cases where it might be hard to judge if a pattern is an error or a genuine pattern in the dataset.

It is also worth noting that our tool is able to reduce disparity between the privileged and the unprivileged group but it may not be able to close the gap entirely. This can be due to missing proxy variables whose influence is unaccounted for or maybe because linear models are too simple to capture the relationship between a node and its parents. Future work might use non-linear SEMs and causal discovery algorithms like Fast Causal Inference (FCI)
%\cite{fci} 
that can better deal with missing attributes.
%One possible reason can be  
%It should be noted that our debiasing We don't get perfect results because some of the proxy variable might be missing or maybe because the relationship between is not accurately captured by linear models.

%A possible next step will be to try other causal discovery algorithms like Fast Causal Inference (FCI)\cite{fci} that can better deal with missing attributes.
% latent confounders
%D-BIAS leverages domain knowledge to refine the causal model generated by the CDA. Future work might incorporate the domain knowledge as a parameter into the CDA to generate better causal model. 
%\textcolor{blue}{How to differentiate between Refining and Debiasing stage? Which errors are caused due to limitations of causal discovery model or due to bias in data? }.  

\textbf{Scalability.} As the size of the dataset increases in terms of features and rows, scalability can become an issue. 
%As the dataset size increases
With dataset size, the time to generate a causal network, debiasing data, finding paths between nodes and computing evaluation metrics will all increase proportionally. Among all these steps, the process to generate the default causal network might be the most computationally expensive. So, future work should employ GPU-based parallel implementation of PC algorithm like cuPC \cite{zarebavani2019cupc} or use inherently faster causal discovery algorithms like F-GES \cite{xie2020visual} to better scale to larger datasets.
%or  that can leverage multiple compute cores simultaneously or employ .
%This can have a negative impact on user experience. 
On the front end, the causal network will become big and complex as the number of features increase. With limited screen space, the user might find it difficult to comprehend the causal network. To alleviate this issue, we have implemented different visual analytics techniques like zooming, panning, filtering weak edges, finding paths, etc.  
%This will enable our tool to deal with larger datasets with minimum lags. 
Future work might optimize graph layout algorithm and explore other visual analytics techniques like node aggregation to help navigate larger graphs better \cite{xie2020visual}.
%\textcolor{blue}{Finding all paths between two nodes in a graph can have exponential complexity.} 
%Future work might also extend support for comparison among multiple groups ($>2$) simultaneously say Blacks, Asians and Hispanics.

%\textbf{Group Selection} D-BIAS supports comparison between two groups at a time as evident from the fourfold display plot, custom subgroup selection panel and fairness metrics. This is in line with the fairness literature which also focuses on comparison between two groups i.e., privileged and underprivileged groups. However, viewing the world from a binary lens might not provide the complete picture. Future work might extend support for comparison among multiple groups simultaneously say black females, white males and black males. 
%Instead of visualizing the difference between groups in terms of some metric like Accuracy, we can visualize individual accuracy numbers for multiple groups (see \cite{fairvis}).   

\textbf{Applications.} In this paper, we have emphasized how our tool can help identify and remove social biases. However, our approach and tool is not limited to social biases. Our tool can incorporate human feedback to realize policy and institutional goals as well (see appendix D.1). For eg., one might strengthen the edge between the nodes \textit{Education} and \textit{Income} to implant a policy intervention where people with higher education are incentivized.
%generate a dataset where people with higher education are incentivized. 
A ML model trained over the resulting dataset is likely to reflect such policy intervention in its predictions.   
%\textcolor{blue}{We saw a glimpse of this in the Synthetic Hiring dataset when Jane strengthened the edge between \textit{Work Experience} and \textit{Job}. Such a change is imposing a policy decision where candidates with higher work experience are incentivized.}
%Furthermore, our tool can be easily extended to generate fair and representative synthetic datasets by drawing exogenous variables from user-specified specific distribution instead of the original dataset. Such datasets can be used to train ML models, augment existing datasets, etc.  
Next, we plan to extend our HITL methodology to tackle biases in other domains such as word embeddings.  

\textbf{Human factors.} Involving a human in the loop for identifying and debiasing data is a double edged sword. On one hand, it is a key strength of our tool as it provides real world domain knowledge and fosters accountability and trust. On the other hand, it can also be its main weakness if the human operating this tool intentionally/unconsciously injects social biases. 
A user can misuse the system in two ways. Firstly, the user can choose to ignore the social biases inherent in the dataset by not acting on the unfair causal edges. Such behaviour renders the system ineffective. Secondly, a biased user can explicitly introduce their own biases in the system by adding/strengthening unfair causal edges. 
%This can be accomplished by adding/strengthening an edge between the sensitive/proxy variable and label variable. 
Since this is a human aided tool, the biases that are inherent to the human user cannot be avoided. Hence, we recommend choosing the user responsibly. 
%A human who is well versed with the domain of the data, sensitivities of the society and who is trusted by the majority of stakeholders might be a good fit. 
%If a user tries to misuse the system, we can always check the system logs and hold the person responsible for their action/inaction. 
Moreover, we can always check the system logs and hold the person responsible for their action/inaction.
%One of the main weaknesses of our system is its vulnerability to the bias of the user. 
  
% If the human is biased, it can cause more bad than good. 

%\textbf{Future Work}
%The next step would be to act on the subjective feedback given by the user study participants. This includes 
%Next step includes adding tooltips for different UI components like fairness metrics, improving graph layout algorithm, visualizing exact changes to the tabular data in the interface itself, etc. We also plan to extend support for more fairness metrics and causal discovery algorithms. Furthermore, it will be interesting to see how different automated debiasing techniques compare against our tool on different evaluation metrics.  
%The user should be free to choose the most relevant fairness metrics from a pool of possible metrics. 
%Lastly, we plan to extend our human in the loop methodology to detect and mitigate bias in different domains like word embeddings, active learning, etc.  
%At what difference in fairness metrics, do people trust automated solution more.
%Compare against set of existing preprocessing techniques

\acknowledgments{
This work was partially funded by NSF grants CNS 1900706, IIS 1527200, IIS 1941613, and NSF SBIR contract 1926949.}

\bibliographystyle{abbrv-doi}

\bibliography{references}

\begin{thebibliography}{10}

\bibitem{fairsight}
Y.~Ahn and Y.-R. Lin.
\newblock Fairsight: Visual analytics for fairness in decision making.
\newblock {\em IEEE Transactions on Visualization and Computer Graphics},
  26(1):1086--1095, 2019.

\bibitem{barocas2017fairness}
S.~Barocas, M.~Hardt, and A.~Narayanan.
\newblock Fairness in machine learning.
\newblock {\em Nips tutorial}, 1:2017, 2017.

\bibitem{inprocess}
Y.~Bechavod and K.~Ligett.
\newblock Learning fair classifiers: A regularization-inspired approach.
\newblock {\em arXiv:1707.00044}, 2017.

\bibitem{bechavod2017penalizing}
Y.~Bechavod and K.~Ligett.
\newblock Penalizing unfairness in binary classification.
\newblock {\em arXiv preprint arXiv:1707.00044}, 2017.

\bibitem{aif360}
R.~K. Bellamy, K.~Dey, M.~Hind, et~al.
\newblock Ai fairness 360: An extensible toolkit for detecting, understanding,
  and mitigating unwanted algorithmic bias.
\newblock {\em arXiv}, 2018.

\bibitem{bickel1975sex}
P.~J. Bickel, E.~A. Hammel, and J.~W. O'Connell.
\newblock Sex bias in graduate admissions: Data from berkeley.
\newblock {\em Science}, 1975.

\bibitem{sus}
J.~Brooke et~al.
\newblock Sus-a quick and dirty usability scale.
\newblock {\em Usability evaluation in industry}, 189(194):4--7, 1996.

\bibitem{fairvis}
{\'A}.~A. Cabrera, W.~Epperson, F.~Hohman, M.~Kahng, J.~Morgenstern, and D.~H.
  Chau.
\newblock Fairvis: Visual analytics for discovering intersectional bias in
  machine learning.
\newblock {\em arXiv}, 2019.

\bibitem{cai2019effects}
C.~J. Cai, J.~Jongejan, and J.~Holbrook.
\newblock The effects of example-based explanations in a machine learning
  interface.
\newblock In {\em Proceedings of the 24th International Conference on
  Intelligent User Interfaces}, 2019.

\bibitem{nipsIBM}
F.~P. Calmon, D.~Wei, K.~N. Ramamurthy, and K.~R. Varshney.
\newblock Optimized data pre-processing for discrimination prevention.
\newblock {\em arXiv preprint arXiv:1704.03354}, 2017.

\bibitem{chiappa2019path}
S.~Chiappa.
\newblock Path-specific counterfactual fairness.
\newblock In {\em Proceedings of the AAAI Conference on Artificial
  Intelligence}, 2019.

\bibitem{chiappa2018causal}
S.~Chiappa and W.~S. Isaac.
\newblock A causal bayesian networks viewpoint on fairness.
\newblock In {\em IFIP International Summer School on Privacy and Identity
  Management}, pp. 3--20. Springer, 2018.

\bibitem{chickering2002optimal}
D.~M. Chickering.
\newblock Optimal structure identification with greedy search.
\newblock {\em Journal of machine learning research}, 2002.

\bibitem{Colombo2014Order-independentLearning}
D.~Colombo and M.~Maathuis.
\newblock {Order-independent constraint-based causal structure learning}.
\newblock {\em Journal of Machine Learning Research}, 2014.

\bibitem{amazonHiringSexist}
J.~Dastin.
\newblock Amazon scraps secret ai recruiting tool that showed bias against
  women.
\newblock Reuters, 2018.

\bibitem{dietvorst2016overcoming}
B.~J. Dietvorst, J.~P. Simmons, and C.~Massey.
\newblock Overcoming algorithm aversion: People will use imperfect algorithms
  if they can (even slightly) modify them.
\newblock {\em Management Science}, 2016.

\bibitem{drozdal2020trust}
J.~Drozdal, J.~Weisz, D.~Wang, G.~Dass, B.~Yao, C.~Zhao, M.~Muller, L.~Ju, and
  H.~Su.
\newblock Trust in automl: exploring information needs for establishing trust
  in automated machine learning systems.
\newblock In {\em International Conference on Intelligent User Interfaces},
  2020.

\bibitem{dwork2018group}
C.~Dwork and C.~Ilvento.
\newblock Group fairness under composition.
\newblock In {\em Conference on Fairness, Accountability, and Transparency
  (FAT* 2018)}.

\bibitem{cytoscape}
M.~Franz, C.~T. Lopes, G.~Huck, Y.~Dong, O.~Sumer, and G.~D. Bader.
\newblock Cytoscape. js: a graph theory library for visualisation and analysis.
\newblock {\em Bioinformatics}, 32(2):309--311, 2016.

\bibitem{friendly1994fourfold}
M.~Friendly.
\newblock A fourfold display for 2 by 2 by k tables.
\newblock Technical report, Citeseer, 1994.

\bibitem{ghai2021wordbias}
B.~Ghai, M.~N. Hoque, and K.~Mueller.
\newblock Wordbias: An interactive visual tool for discovering intersectional
  biases encoded in word embeddings.
\newblock In {\em ACM CHI, Extended Abstracts}, 2021.

\bibitem{XAL}
B.~Ghai, Q.~V. Liao, Y.~Zhang, R.~Bellamy, and K.~Mueller.
\newblock Explainable active learning (xal) toward ai explanations as
  interfaces for machine teachers.
\newblock {\em Proc. ACM CSCW}, 2021.

\bibitem{ghai2022cascaded}
B.~Ghai, M.~Mishra, and K.~Mueller.
\newblock Cascaded debiasing: Studying the cumulative effect of multiple
  fairness-enhancing interventions.
\newblock {\em arXiv preprint arXiv:2202.03734}, 2022.

\bibitem{glymour2019review}
C.~Glymour, K.~Zhang, and P.~Spirtes.
\newblock Review of causal discovery methods based on graphical models.
\newblock {\em Frontiers in genetics}, 2019.

\bibitem{gower}
J.~C. Gower.
\newblock A general coefficient of similarity and some of its properties.
\newblock {\em Biometrics}, pp. 857--871, 1971.

\bibitem{hajian2013dataModification}
S.~Hajian and J.~Domingo-Ferrer.
\newblock A methodology for direct and indirect discrimination prevention in
  data mining.
\newblock {\em IEEE transactions on knowledge and data engineering}, 2013.

\bibitem{nasa_tlx}
S.~G. Hart and L.~E. Staveland.
\newblock Development of nasa-tlx (task load index): Results of empirical and
  theoretical research.
\newblock In {\em Advances in Psychology}, vol.~52, pp. 139--183. 1988.

\bibitem{hoque2021outcome}
M.~N. Hoque and K.~Mueller.
\newblock Outcome-explorer: A causality guided interactive visual interface for
  interpretable algorithmic decision making.
\newblock {\em arXiv preprint arXiv:2101.00633}, 2021.

\bibitem{cdt}
D.~Kalainathan and O.~Goudet.
\newblock Causal discovery toolbox: Uncover causal relationships in python.
\newblock {\em arXiv preprint arXiv:1903.02278}, 2019.

\bibitem{pcalg}
M.~Kalisch, M.~M\"achler, D.~Colombo, M.~H. Maathuis, and P.~B\"uhlmann.
\newblock Causal inference using graphical models with the {R} package {pcalg}.
\newblock {\em Journal of Statistical Software}, 47(11):1--26, 2012.

\bibitem{kamiran2009classifying}
F.~Kamiran and T.~Calders.
\newblock Classifying without discriminating.
\newblock In {\em Computer, Control and Communication, 2009. IC4 2009. 2nd
  International Conference on}, pp. 1--6. IEEE, 2009.

\bibitem{kamiran2012}
F.~Kamiran and T.~Calders.
\newblock Data preprocessing techniques for classification without
  discrimination.
\newblock {\em Knowledge and Information Systems}, 33(1):1--33, 2012.

\bibitem{kleinberg2016inherent}
J.~Kleinberg, S.~Mullainathan, and M.~Raghavan.
\newblock Inherent trade-offs in the fair determination of risk scores.
\newblock {\em arXiv}, 2016.

\bibitem{kusner2017counterfactual}
M.~J. Kusner, J.~Loftus, C.~Russell, and R.~Silva.
\newblock Counterfactual fairness.
\newblock In {\em Advances in Neural Information Processing Systems}.

\bibitem{MXM}
V.~Lagani, G.~Athineou, A.~Farcomeni, M.~Tsagris, and I.~Tsamardinos.
\newblock Feature selection with the {R} package {MXM}: Discovering
  statistically equivalent feature subsets.
\newblock {\em Journal of Statistical Software}, 80(7), 2017. doi: {{%
10\hspace{.1pt}\discretionary{.}{%
}{.}\hspace{.4pt}18637\discretionary{/}{%
}{/}jss\hspace{.1pt}\discretionary{.}{%
}{.}\hspace{.4pt}v080\hspace{.1pt}\discretionary{.}{%
}{.}\hspace{.4pt}i07}}


\bibitem{lohia2018bias}
P.~K. Lohia, K.~N. Ramamurthy, M.~Bhide, D.~Saha, K.~R. Varshney, and R.~Puri.
\newblock Bias mitigation post-processing for individual and group fairness.
\newblock {\em arXiv preprint arXiv:1812.06135}, 2018.

\bibitem{madsen2000measuring}
M.~Madsen and S.~Gregor.
\newblock Measuring human-computer trust.
\newblock In {\em 11th australasian conference on information systems}, 2000.

\bibitem{trust1}
D.~H. McKnight, V.~Choudhury, and C.~Kacmar.
\newblock Developing and validating trust measures for e-commerce: An
  integrative typology.
\newblock {\em Information Systems Research}, 2002.

\bibitem{trust2}
D.~H. McKnight, L.~L. Cummings, and N.~L. Chervany.
\newblock Initial trust formation in new organizational relationships.
\newblock {\em Academy of Management Review}, 23(3):473--490, 1998.

\bibitem{arvindTalk}
A.~Narayanan.
\newblock 21 fairness definitions and their politics.
\newblock Conference on Fairness, Accountability, and Transparency, NYC, 2018.

\bibitem{Pearl2000}
J.~Pearl.
\newblock {\em {Causality: Models, Reasoning, and Inference}}.
\newblock Cambridge University Press, 2000.

\bibitem{Pearl2009}
J.~Pearl.
\newblock {Causal inference in statistics: An overview}.
\newblock {\em Statistics Surveys}, 3(0):96--146, 2009. doi: {{%
10\hspace{.1pt}\discretionary{.}{%
}{.}\hspace{.4pt}1214\discretionary{/}{%
}{/}09\discretionary{%
}{-}{-}SS057}}


\bibitem{rajabi2021tabfairgan}
A.~Rajabi and O.~O. Garibay.
\newblock Tabfairgan: Fair tabular data generation with generative adversarial
  networks, 2021.

\bibitem{fges}
J.~Ramsey, M.~Glymour, R.~Sanchez-Romero, and C.~Glymour.
\newblock A million variables and more: the fast greedy equivalence search
  algorithm for learning high-dimensional graphical causal models, with an
  application to functional magnetic resonance images.
\newblock {\em International journal of data science and analytics}, 2017.

\bibitem{challenges}
X.~Shen, S.~Ma, P.~Vemuri, and G.~Simon.
\newblock Challenges and opportunities with causal discovery algorithms:
  Application to alzheimer’s pathophysiology.
\newblock {\em Scientific Reports}, 2020.

\bibitem{sofrygin2017simcausal}
O.~Sofrygin, M.~J. van~der Laan, and R.~Neugebauer.
\newblock Simcausal r package: conducting transparent and reproducible
  simulation studies of causal effect estimation with complex longitudinal
  data.
\newblock {\em J. of Statistical Software}, 2017.

\bibitem{Spirtes2000}
P.~Spirtes, C.~Glymour, and R.~Scheines.
\newblock {\em {Causation, Prediction, and Search}}.
\newblock MIT Press, Cambridge, MA, 2000.

\bibitem{tsagris2018constraint}
M.~Tsagris, G.~Borboudakis, V.~Lagani, and I.~Tsamardinos.
\newblock Constraint-based causal discovery with mixed data.
\newblock {\em International journal of data science and analytics}, 2018.

\bibitem{wachter2021fairness}
S.~Wachter, B.~Mittelstadt, and C.~Russell.
\newblock Why fairness cannot be automated: Bridging the gap between eu
  non-discrimination law and ai.
\newblock {\em Computer Law \& Security Review}, 41:105567, 2021.

\bibitem{Wang2016}
J.~Wang and K.~Mueller.
\newblock {The Visual Causality Analyst: An Interactive Interface for Causal
  Reasoning}.
\newblock {\em IEEE Transactions on Visualization and Computer Graphics}, 2016.
  doi: {{%
10\hspace{.1pt}\discretionary{.}{%
}{.}\hspace{.4pt}1109\discretionary{/}{%
}{/}TVCG\hspace{.1pt}\discretionary{.}{%
}{.}\hspace{.4pt}2015\hspace{.1pt}\discretionary{.}{%
}{.}\hspace{.4pt}2467931}}


\bibitem{wang2017visual}
J.~Wang and K.~Mueller.
\newblock Visual causality analysis made practical.
\newblock In {\em Proc. IEEE VAST}, pp. 151--161, 2017.

\bibitem{DiscriLens}
Q.~Wang, Z.~Xu, Z.~Chen, Y.~Wang, S.~Liu, and H.~Qu.
\newblock Visual analysis of discrimination in machine learning.
\newblock {\em arXiv}, 2020.

\bibitem{wu2018discrimination}
Y.~Wu, L.~Zhang, and X.~Wu.
\newblock On discrimination discovery and removal in ranked data using causal
  graph.
\newblock In {\em ACM SIGKDD Conference on Knowledge Discovery \& Data Mining},
  2018.

\bibitem{xie2021fairrankvis}
T.~Xie, Y.~Ma, J.~Kang, H.~Tong, and R.~Maciejewski.
\newblock Fairrankvis: A visual analytics framework for exploring algorithmic
  fairness in graph mining models.
\newblock {\em IEEE Transactions on Visualization and Computer Graphics},
  28(1):368--377, 2021.

\bibitem{xie2020visual}
X.~Xie, F.~Du, and Y.~Wu.
\newblock A visual analytics approach for exploratory causal analysis:
  Exploration, validation, and applications.
\newblock {\em IEEE Transactions on Visualization and Computer Graphics},
  27(2):1448--1458, 2020.

\bibitem{silva}
J.~N. Yan, Z.~Gu, H.~Lin, and J.~M. Rzeszotarski.
\newblock Silva: Interactively assessing machine learning fairness using
  causality.
\newblock In {\em ACM CHI Conference on Human Factors in Computing Systems},
  2020.

\bibitem{zarebavani2019cupc}
B.~Zarebavani, F.~Jafarinejad, M.~Hashemi, and S.~Salehkaleybar.
\newblock cupc: Cuda-based parallel pc algorithm for causal structure learning
  on gpu.
\newblock {\em IEEE Trans. on Parallel and Distributed Systems},
  31(3):530--542, 2019.

\bibitem{zemel2013learning}
R.~Zemel, Y.~Wu, K.~Swersky, T.~Pitassi, and C.~Dwork.
\newblock Learning fair representations.
\newblock In {\em Intern. Conf. on Machine Learning}, pp. 325--333, 2013.

\bibitem{zhang2017anti}
L.~Zhang and X.~Wu.
\newblock Anti-discrimination learning: a causal modeling-based framework.
\newblock {\em Intern. J. Data Science and Analytics}, 4(1):1--16, 2017.

\bibitem{zhang2017causal}
L.~Zhang, Y.~Wu, and X.~Wu.
\newblock A causal framework for discovering and removing direct and indirect
  discrimination.
\newblock In {\em Proceedings of the 26th International Joint Conference on
  Artificial Intelligence}, 2017.

\end{thebibliography}

\clearpage

\begin{comment}
\begin{table*}[h!]
\centering
%\vspace{10em}
\begin{tabular}{c c c} 
  & \huge\textbf{APPENDIX} &  \\ 
\end{tabular}
%\vspace{15em}
\end{table*}
\end{comment}

\appendix
%\makebox[2\linewidth]{Get a stretcher}
\begin{center}
    \huge\textbf{APPENDIX}\\
\end{center}
\vspace{2em}

%\hspace*{\fill}
%\makebox[\linewidth]{\textbf{\huge APPENDIX}} %\hspace*{\fill}
\section{Implementation}
D-BIAS is implemented as a web application built over python based web framework \textit{Flask}. To generate the causal network, we used two R based libraries, i.e., pcalg \cite{pcalg} for the PC Algorithm and \textit{MXM} \cite{MXM} for the conditional independence test. For accessing these R based libraries in python, we used the open source python framework \textit{cdt} \cite{cdt} and made some modifications to suit our needs. For the front end, we have used javascript based open source graph library \textit{Cytoscape.js} \cite{cytoscape} for rendering the causal network. Additionally, we used libraries like \textit{D3.js}, \textit{Plotly.js}, \textit{Bootstrap}, \textit{noUiSlider}, etc. for rendering different visual components. 
%The source code for this project can be accessed at %\url{bit.ly/3gRGJgA}.
%\url{github.com/bhavyaghai/D-BIAS}.
%It will be made publicly available upon acceptance for easy reproducibility.

%\section{Background}
%In this section, we provide a brief background about causal discovery and structural equation model, and how our work relates to each of them.

\section{Causal Discovery}
%In this section, we provide a brief background about causal discovery and how our work relates to each of them
Randomized controlled trials (RCTs) have been used across disciplines to establish causal relations \cite{glymour2019review}. However, conducting such experiments can cost a lot of money, time or might simply be impossible \cite{glymour2019review}. Moreover, in this work, we are dealing with observational data which renders RCTs ineffective. Hence, we have relied on a popular alternative, namely Causal Discovery Algorithms (CDAs). 
Given observational data, CDAs find causal relationships among different attributes in the dataset \cite{glymour2019review}.
%Causal Discovery Algorithms (CDA) help discover causal relations from pure observational datasets.  
Causal relations returned by CDAs can be graphically represented via a directed acyclic graph (DAG), also known as causal graph, where each node represents a data attribute and each edge represents a causal relation. For example, a directed edge from X to Y signify that X is the cause of Y, i.e., a change in X will cause a change in Y. The set of causal discovery algorithms can be broadly classified into constraint based methods such as PC \cite{Spirtes2000}, Fast Causal Inference (FCI), etc., and score based algorithms such as Greedy Equivalence Search (GES) \cite{chickering2002optimal}, Fast GES \cite{fges}, etc. Among the set of possible causal DAGs, constraint based CDAs rely on clever schedule of conditional independence tests whereas score based CDAs rely on some fitness metric like BIC score to filter to the final causal graph. 
%Constraint based CDAs are widely applicable as they can handle different types of data distributions. 
%They rely on conditional independence tests to generate the causal graph whereas score based CDAs generates a set of graphs and then compare them based on some metric like Bayes Information criterion to get to the final causal graph. 

In this work, we have used PC algorithm~\cite{Spirtes2000} 
%(constraint based CDA)
as it can handle mixed data types and provide asymptotically correct results. PC algorithm, named after its inventors \underline{P}eter Spirtes and \underline{C}lark Glymour, is a constraint based CDA. PC algorithm starts with a complete undirected graph where each node is an attribute in the tabular dataset. Thereafter, it starts filtering edges based on conditional independence tests. For example, if nodes X and Y are unconditionally independent, then the edge X $\to$ Y  is removed. Here, the tabular dataset can consist of numeric and categorical attributes so we have used the \textit{Symmetric Conditional Independence Test} which can deal with mixed data types \cite{tsagris2018constraint}. After filtering, it uses a set of orientation rules to direct edges. Each directed edge represents a causal relationship from cause to effect. PC algorithm returns a partially directed acyclic graph, i.e., a DAG with some undirected edges.

It should be noted that causal discovery algorithms operate under a set of assumptions and have their own limitations. For example, PC algorithm assumes that no confounders (direct common cause of two variables) are missing from the dataset. In a practical setting, such factors can lead to an imperfect causal graph. One way to refine such causal graphs is by injecting domain knowledge. In line with existing work \cite{Wang2016, wang2017visual, hoque2021outcome}, we have employed a visual analytics approach to augment CDAs with human knowledge to get to the true causal graph. 

\begin{figure}[tb]
 \centering 
 \includegraphics[width=0.9\columnwidth]{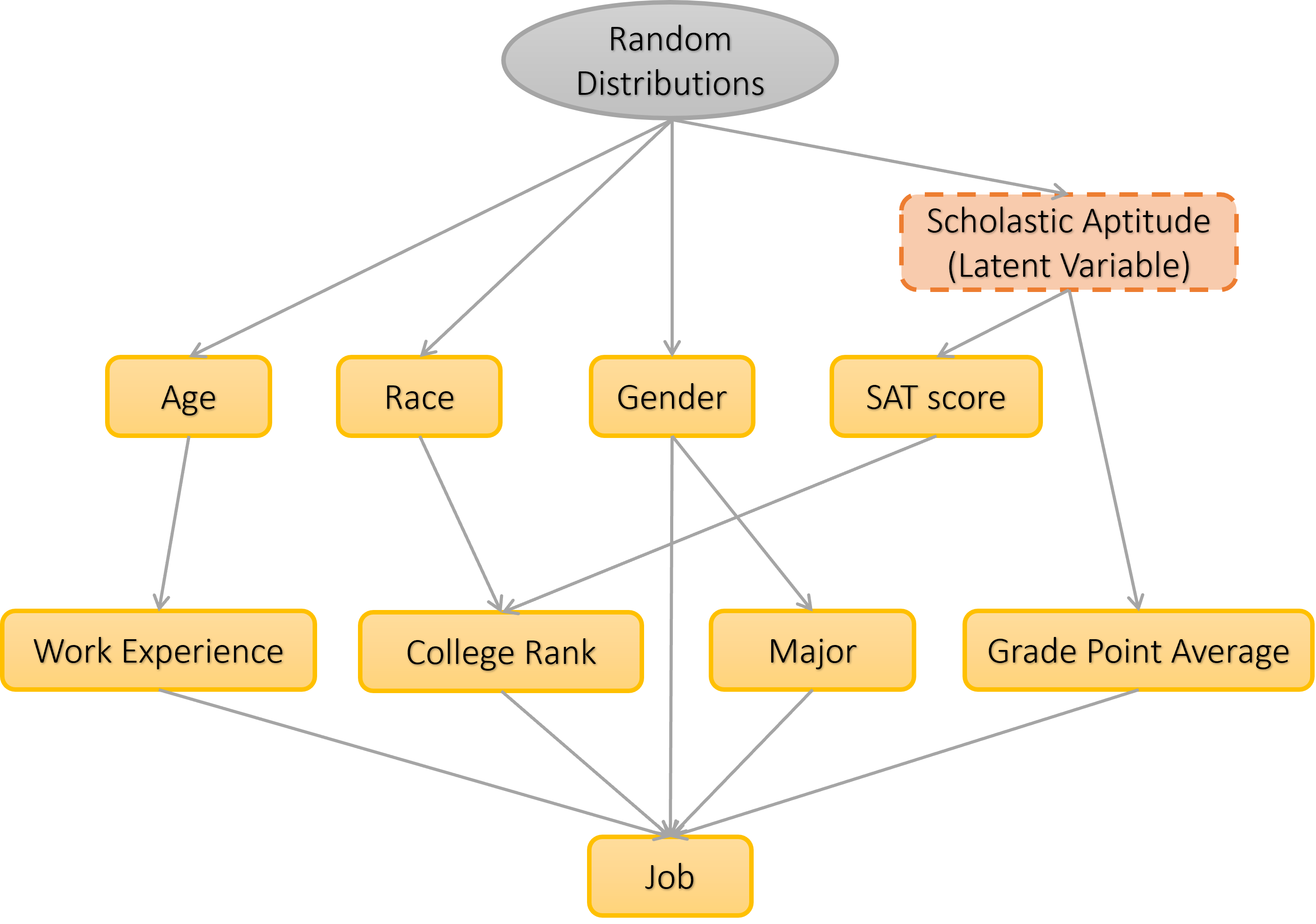}
 \caption{Data generating mechanism for the synthetic hiring dataset. Each node, colored in gold, represents a feature in the dataset; each arrow represents a causal relation between features.}  
 \label{fig:syn_data_gen}
\end{figure}

%In our case, we are dealing with observational data which renders this method ineffective.
%However, conducting such experiments can cost a lot of money, time or might simply be impossible\cite{glymour2019review}. 
%Given a tabular dataset, CDA helps identify causal relations between different attributes in the dataset \cite{glymour2019review}. 
%In our case, we need to discover causal relations from observational data which renders the prior method ineffective. 
% In general how do causal discovery work
%Another data driven method of determining causal relations using purely observational data is called causal discovery.   
%There are many variants of PC algorithm. 
%We have used PC Stable algorithm. 

%One of the ways to represent such causal relations is via directed acyclic graphs. It consists of a set of nodes and edges where each node represents an attribute in the data and each edge represents a causal relationship. 

\begin{figure*}[t]
 \centering 
 \includegraphics[width=2\columnwidth]{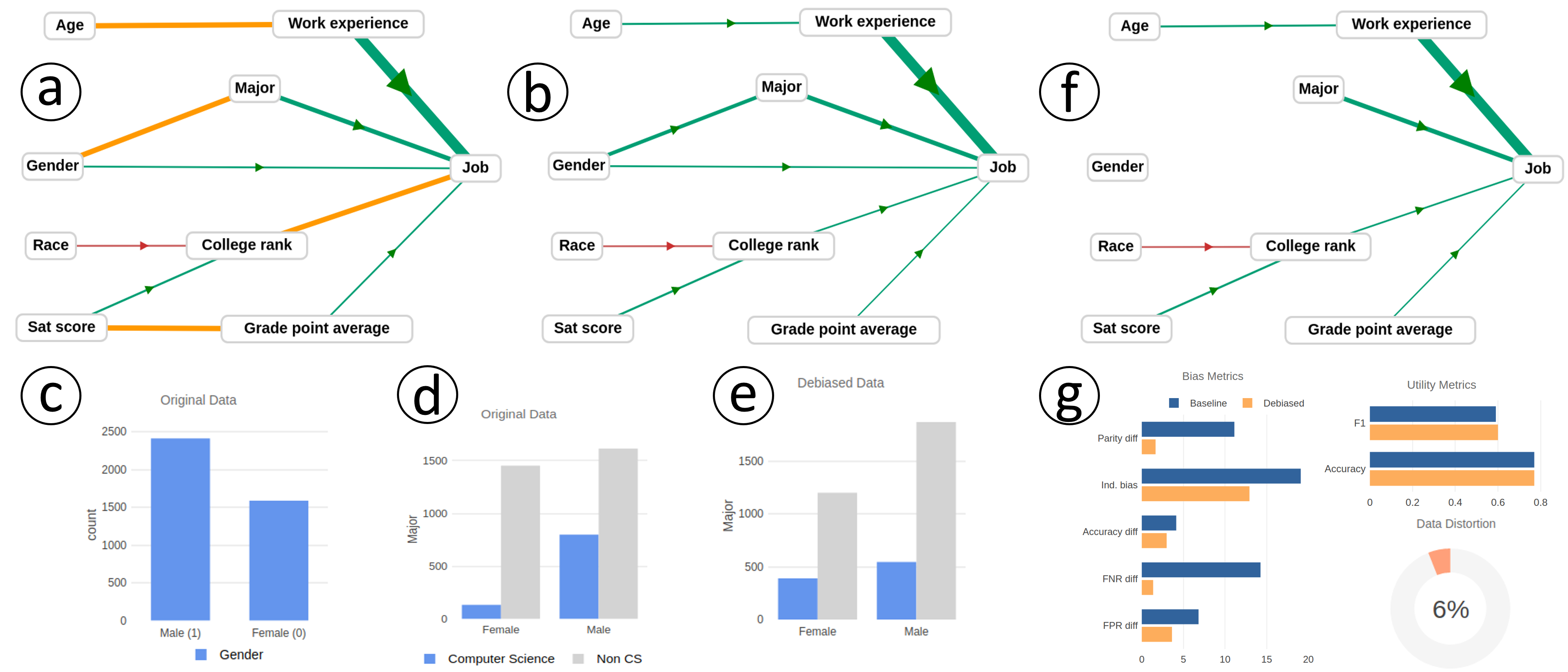}

 \caption{Different steps for the case study of the Synthetic Hiring dataset (a) Causal model generated using automated techniques (b) Refined causal model (c) Clicking the Gender node visualizes its distribution as a bar chart (d) Bivariate relation between Major and Gender for the original dataset (e) Bivariate relation between Major and Gender for the debiased dataset (f) Debiased causal model (g) Evaluation metrics to compare our results against the baseline debiasing approach.} 
 \label{fig:synthetic}
\end{figure*}

\begin{figure*}[h!]
 \centering 
 \includegraphics[width=2\columnwidth]{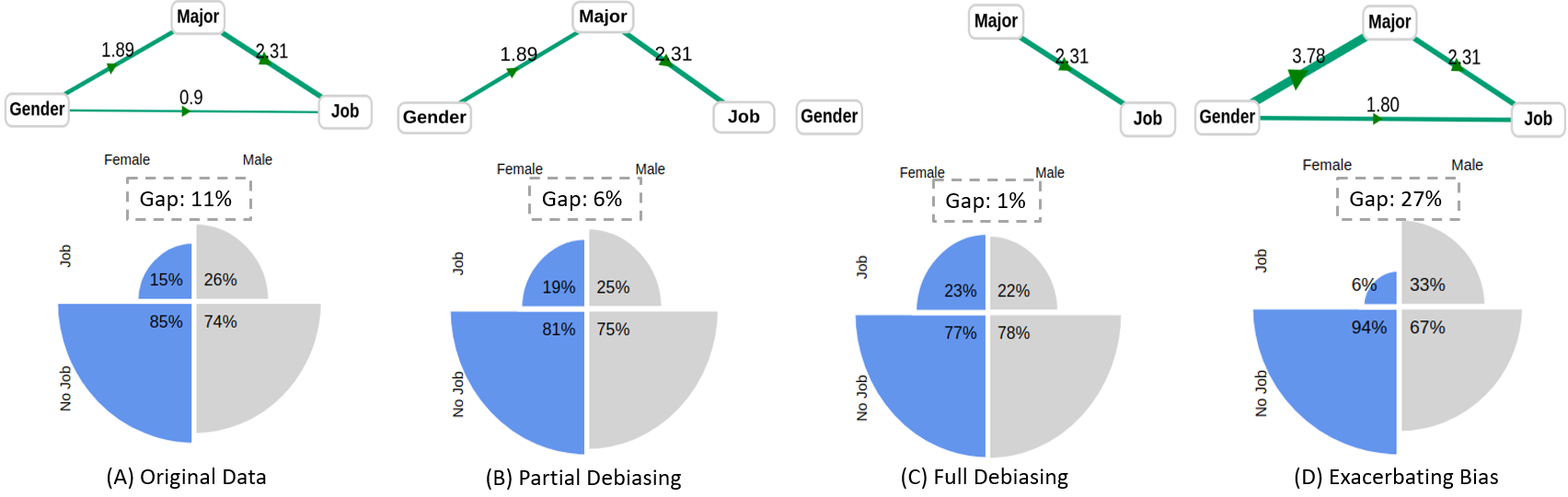}

 \caption{Fourfold displays corresponding to different user interactions for the synthetic hiring dataset. Here, we have we have shown a subset of the causal network which connects the Gender node with the Job node. The left half of each fourfold display represents females and the right half represents males. The top left quadrant represents the percentage of females who got the job, the top right quadrant represents the percentage of males who got the job, and so on.} 
 \label{fig:syn_fourfold}
\end{figure*}

\section{Synthetic Hiring Dataset}
We generated a synthetic hiring dataset fraught with gender and racial bias to test our tool.
%\footnote{Refer to supplementary material to access this dataset.}. 
Unlike real world datasets, we exactly know the underlying data generation process for a synthetic dataset. This helps better gauge the overall performance of our tool.

%we can easily compare use a synthetic dataset instead of a real dataset for our first demonstration because the source of the bias can be clearly explained using a relatively small number of attribute variables (real world examples will be presented in \autoref{sec:usage_scenario}). 
This dataset consists of 4,000 rows and 9 columns (3 numeric and 6 categorical). Each row represents a job candidate with features like gender, work experience, age, etc. The output variable `Job' represents whether a candidate got the job or not. 
%We started by generating gender, race, age and scholastic aptitude from random distributions. Here, \textit{Scholastic aptitude} is a latent variable so it can impact other features but it won't be part of the dataset. Thereafter, we generate the rest of the features as a weighted sum of their parent nodes and random noise.
We generated the exogenous variables such as race, gender, age, etc. by drawing random samples from different distributions. Here, we used binomial distribution for categorical variables and uniform distribution for numeric variables. Thereafter, each endogenous variable is generated as a linear function of its parent nodes.  
%For categorical variables such as gender and race, we used binomial distributions; for continuous variables such as , eg., we have binomial distribution for 
The inter-dependencies between all features is represented in \autoref{fig:syn_data_gen}. In an effort to mimic the real world challenges, we introduced sampling bias for the gender variable such that there are more males than females. Moreover, we introduced a latent confounding variable (scholastic aptitude), i.e., a relevant variable that is missing from the dataset. 

%Next, we will demonstrate the workflow and functioning of our tool using this synthetic dataset. Later, we will also test our tool on a real world dataset (see \autoref{sec:use_case1}) 

%\begin{figure*}[tb]
% \centering 
% \includegraphics[width=\columnwidth]{figure/synthetic_usage_scenario.png}
%  \setlength{\belowcaptionskip}{-8pt}
%\setlength{\abovecaptionskip}{-4pt}
% \caption{Usage Scenario: Synthetic Hiring dataset (a) Initial causal network returned by the PC Algorithm (b) Unequal representation of females in the dataset (c) Refined causal network (d) Relationship between Major and Job (e) Gender debiased causal network (f) Fairness metrics with respect to Gender (g) Gender and Race debiased causal network (h) Fairness metrics with respect to Race}  
% \label{fig:usage_scenario}
%\end{figure*}

\section{Case Study}
In this section, we demonstrate the efficacy of our tool in examining and mitigating algorithmic bias for two other datasets, i.e., synthetic hiring dataset and COMPAS dataset.

\subsection{Hiring Dataset}
Jane works as a Data Scientist for a big software company. Her company receives an abundance  of job applications each day. It is virtually impossible to manually go through each of those applications, so she's tasked with building an AI tool that can filter out weak applications. Learning from Amazon's example \cite{amazonHiringSexist}, she's aware that such recruiting tools can be discriminatory towards minorities and pose serious challenges for her organization. Hence, she decides to use D-BIAS to check if there exists some bias in the training data, and if so, mitigate its effects by debiasing the data before moving forward with the tool.
%Jane knows that her company's hiring policy emphasizes grade point average, work experience, and the college rank of the applicant and does not discriminate on the basis of race or sex of the applicant. However, after reading how bias can creep into seemingly unbiased decision making processes she decides to use D-BIAS to analyze her dataset. She identifies $Race$ and $Sex$ as the two protected variables in the dataset. She would like to ensure that there is no discrimination based on these attributes. In the event that there is some bias in the hiring process of the company she would like to mitigate the effect of the bias.% so that the hiring process does not discriminate on the basis of race or sex. 

\textbf{Generating the causal network.} Jane uploads the synthetic hiring dataset into the system which contains records of past applicants profiles and whether they got the job. She selects \textit{Job} as the label variable and chooses \textit{Gender}, \textit{Race}, \textit{Major}, \textit{Grade Point Average}, \textit{College Rank}, \textit{Job} as the nominal attributes. Next, she clicks on the ``Causal Model" button to generate the causal network with p=0.01 (see \autoref{fig:synthetic}(a)). As we are dealing with a synthetic dataset, we can compare the auto-generated causal network using PC algorithm with the ground truth (see \autoref{fig:syn_data_gen}). On comparison, we observe that the automated causal discovery algorithm was able to correctly identify many causal relations such as \textit{Work experience} $\to$ \textit{Job} and \textit{SAT score} $\to$ \textit{College rank}; there are other causal relations which were correctly identified but whose direction could not be determined like \textit{Age} $\to$ \textit{Work experience}. Moreover, there is a wrongly identified causal relation, between \textit{SAT score} and \textit{Grade point average}, that might due to the missing attribute (Scholastic Aptitude). Jane assesses different causal edges and decides to direct the undirected edges. 
%She knows that a cause always precedes its effect. 
She leverages her domain knowledge to direct edges like \textit{Gender} $\to$ \textit{Major}, \textit{College Rank} $\to$ \textit{Job}, etc. to get to the refined causal model (see \autoref{fig:synthetic} (b)).
%refines it as per her domain knowledge.  
%Thereafter, she goes onto refining the causal network by directing undirected edges. 
%One of the guiding principles to determine the direction of an causal edge is that the 
%She directs edges   

\begin{figure*}[t]
 \centering 
 \includegraphics[width=2\columnwidth]{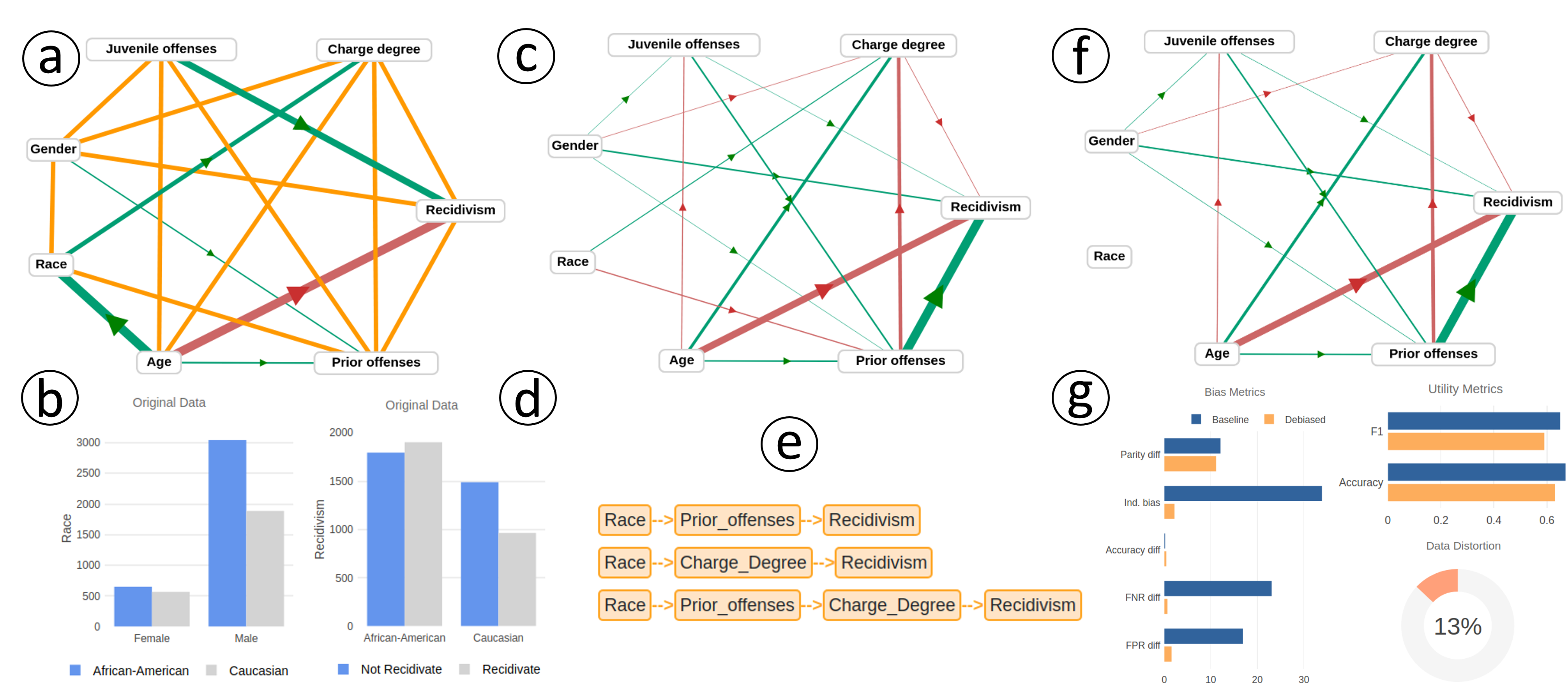}
 \caption{Use case: COMPAS dataset (a) Initial causal network returned by the PC Algorithm (b) Bivariate relationship between Race and Gender (c) Refined causal network (d) Grouped bar chart between Race and Recidivism showing the inherent racial bias (e) All paths between Race and Recidivism (f) Debiased causal network (g) Evaluation metrics to compare our results against the baseline debiasing approach. }  
 \label{fig:compas}
\end{figure*}

\textbf{Auditing for social biases.}
Post-refining, she starts exploring the dataset by clicking on different nodes and edges in the causal network to see their underlying distributions. On clicking the \textit{Gender} node, she finds that there are far more males than females (2412 vs 1588) in the dataset (see \autoref{fig:synthetic}(c)). Such differences in representation can also lead to algorithmic bias. 
%This observation hints towards the possible presence of gender bias. 
On further exploration, she finds two causal paths connecting the gender node with the job node. She is surprised to find that one of those paths is a direct link between the two nodes. This shows that her company's hiring decisions in the past have been directly influenced by gender. The other path connects the two nodes via the major node (see \autoref{fig:synthetic}(b)). Overall, these two causal paths confirms the presence gender bias and gender based disparities inherent in the dataset.  
%represent disparate treatment and disparate impact with respect to gender, respectively. 
To determine the extent of gender bias, she clicks the ``Evaluate Metrics" button from the evaluation panel. As shown in \autoref{fig:syn_fourfold}(A), she observes a 11\% gap in the likelihood of getting a job between the two genders; 26\% of males in the dataset got the job versus 15\% for females. To ensure that the AI tool does not learn such historical biases, she decides to debias the dataset first before training the ML model. 

\textbf{Mitigating bias} 
She starts the debiasing process by flipping the stage toggle on the top panel from \textit{Refine} to
\textit{Debias}. To mitigate the impact of direct gender discrimination, she deletes the edge \textit{Gender} $\to$ \textit{Job}. On evaluation, this results in a significant reduction of bias across all fairness metrics, a small loss in accuracy (1\%), and a slight increase in data distortion (2\%). As shown in \autoref{fig:syn_fourfold}(B), the percentage of females with job increased from 15\% to 19\% while the percentage of males with job decreased from 26\% to 25\%. In totality, the disparity between genders reduced from 11\% to 6\%. Given the minimal loss in utility and the scope of mitigating bias further, Jane decides to continue with the debiasing process. She focuses on the indirect path from \textit{Gender} to \textit{Job} via \textit{Major} node. She selects the edge \textit{Gender} $\to$ \textit{Major} to see their bivariate distribution. As shown in see \autoref{fig:synthetic}(d), she finds that the female representation in computer science is much lower than that of males which is indirectly contributing to bias against women in hiring. This is not something that her organization is directly responsible for. However, she does not want the ML model to learn such a socially undesirable pattern that might eventually lead to gender discrimination. So, she deletes this edge to reach to the debiased causal model (see \autoref{fig:synthetic}(f)). Post-deletion, she focuses on the comparison view to understand the impact of her last intervention. \autoref{fig:synthetic}(d) and \autoref{fig:synthetic}(e) shows the bivariate distribution between gender and major in the original and debiased dataset, respectively. She finds that the number of females in computer science has increased from 135 to 389 while the number of males in computer science has decreased from 799 to 545. Overall, this intervention reduced the disparity between females and males opting for computer science.     
%In a way, this intervention tries to re-imagine history where gender has no influence on college major.   
%It is a much broader problem with many contributing factors. So, she has a choice to keep or remove the edge. In case, she deletes the edge, there will be a sharp decline in utility ($>30\%$) and rise in distortion ($6\%$) for even better fairness metrics (see \autoref{fig:teaser}). Let's say she decides to retain that edge to preserve accuracy. Now, there are no other paths between $Gender$ and $Job$ so she goes on to explore the causal graph again. 
%This time she finds that $Race$ has a indirect path to $Job$ via $College Rank$. On further exploration, she finds that African Americans are not equally represented at elite universities which is indirectly causing racial bias.  Again, she has a choice to make. Let's say this time she decides to remove the edge $race \to College\:Rank$ (see \autoref{fig:usage_scenario}(g)). 

Finally, she re-evaluates the net impact of her interventions as captured by different evaluation metrics. She finds that the disparity between gender has reduced to just 1\% (as shown in \autoref{fig:syn_fourfold} (C)); data distortion has increased to 6\%; utility metrics are virtually the same, and fairness increases across the board as captured by all the 5 fairness metrics (see \autoref{fig:synthetic}(g)). Now, she can simply download the debiased data and use it to train the AI tool for fairer predictions. 
%This will be a much better solution than using the original data in terms of fairness.
In this case study, we have limited ourselves to mitigating gender bias. However, similar process can be followed for mitigating racial bias.  

\textbf{Incorporating institutional goals.} D-BIAS facilitates incorporation of human prior in the system. Human prior is not just limited to social biases. It can be used to implement policy decisions or other institutional goals as well. Let's say Jane's organization changes its hiring policies and now looks for candidates with much higher work experience than before. One way to accomplish this task is by building a custom ML algorithm which incorporates this objective. Another way this can be accomplished is by modifying the dataset such that candidates with higher experience get the Job. Jane can accomplish this objective by simply selecting the edge \textit{Work Experience} $\to$ \textit{Job} and strengthening that edge by some percentage points depending on the policy. This will ensure that the debiased data inherits this policy. Now, Jane can train any vanilla ML algorithm over the debiased dataset to accomplish this goal.
%We ask the user to manually select the nominal attributes in the data because it is difficult to differentiate between nominal and ordinal attributes automatically. 

It is important to note that the flexibility offered by the D-BIAS tool can also be leveraged by malicious users to exacerbate bias. For eg., one might strengthen the edges \textit{Gender} $\to$ \textit{Job} and \textit{Gender} $\to$ \textit{Major} by a 100\% (as shown in the \autoref{fig:syn_fourfold}(D)). This will lead to a surge in disparity between males and females from 11\% to 27\%.

\subsection{COMPAS dataset}
\label{sec:use_case1}

COMPAS is a popular dataset often used in the fairness literature \cite{aif360}. It pertains to the criminal defendants from Broward County, Florida. The task is to predict recidivism within the next 2 years. In other words, we are to classify if a convicted criminal will reoffend in the next 2 years. After prepossessing, we ended up with 6,150 rows and 7 columns. Here, each row corresponds to an individual described by attributes such as number of juvenile offenses, charge degree (felony, misdemeanor), race (Caucasian, African-American), age, gender, etc.  The sensitive attribute is race. In the following, we will demonstrate how D-BIAS can help identify and alleviate racial bias from this dataset. 
%\textcolor{blue}{maybe use different dataset}
%\footnote{Refer to \url{https://github.com/bhavyaghai/d_Bias} for the exact procedure and data}.

\begin{figure*}[t]
 \centering 
 \includegraphics[width=2\columnwidth]{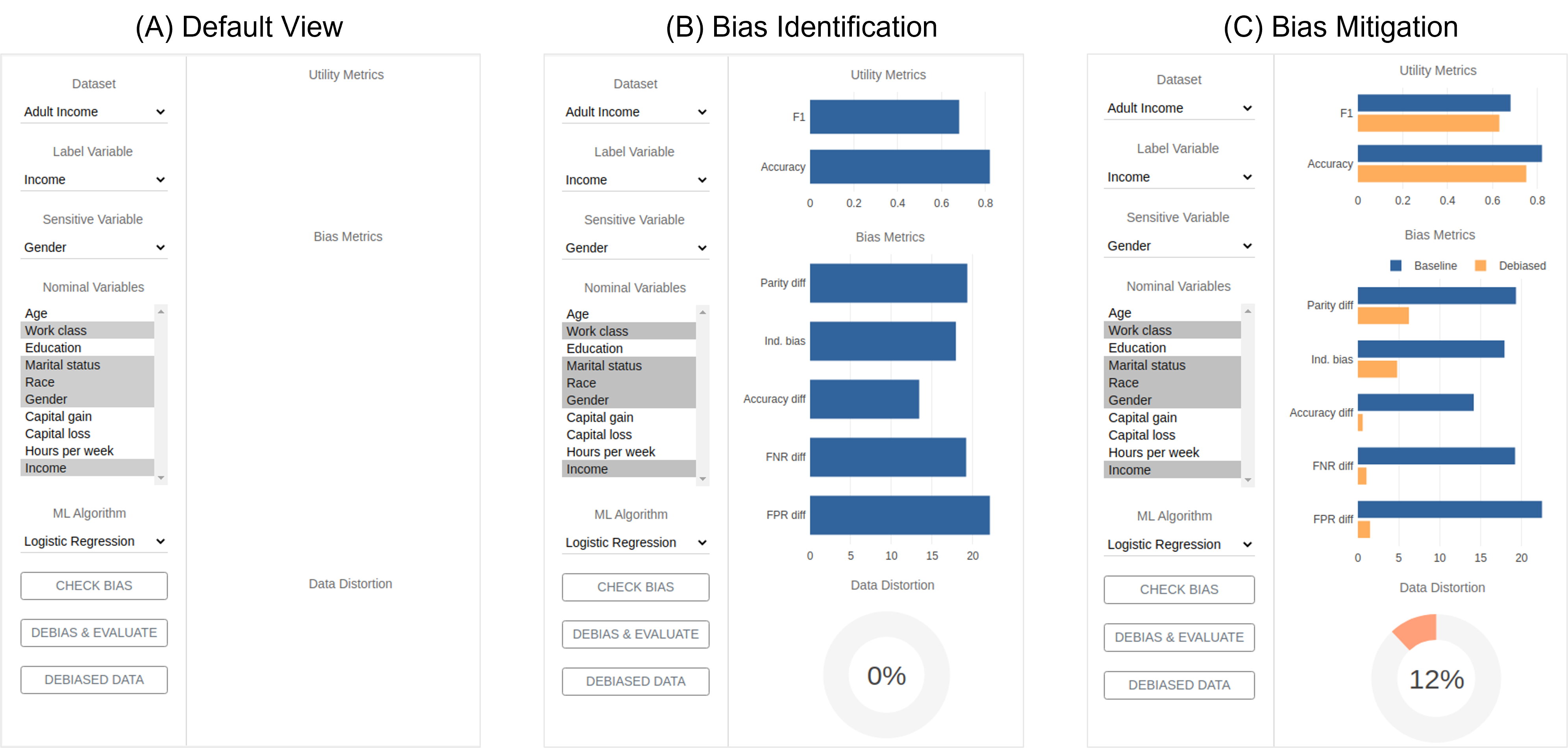}
 \caption{Visual interface of the baseline tool (A) Default view (B) Bias Identification (C) Bias mitigation }  
 \label{fig:baseline_stages}
\end{figure*}

   \textbf{Generating the causal network} We start by selecting COMPAS as the dataset, $Recidivism$ as the label variable, and \textit{Gender}, \textit{Race}, \textit{Recidivism}, \textit{Charge degree} as the nominal variables from the generator panel. Thereafter, we click ``Causal Model" to generate the causal network as shown in \autoref{fig:compas}(a). We observe that there are a lot of undirected edges (in orange). This can be due to missing attributes or due to the limitation of the PC algorithm in determining their direction. We go through all these edges and deal with them based on our domain knowledge. For example, we direct edges such as \textit{Prior Offences} $\to$ \textit{Recidivism}, \textit{Charge degree} $\to$ \textit{Recidivism}, etc. We direct these edges in this way since such a relationship is more plausible than the contrary. Interestingly, we observe an undirected edge between \textit{Gender} and \textit{Race}, and a directed edge between \textit{Age} and \textit{Race}. From our domain knowledge, we know that these relationships do not exist in the real world. To investigate further, we click on these edges to see their corresponding distributions. We observed a sampling bias for both the cases that might have led to the false detection of these edges. As shown in \autoref{fig:compas}(b), we observe that there are a lot more males than females in the dataset. Also, there are more african-americans than caucasians for each gender. This observation hints the need to have a more diverse dataset with respect to gender and race. However, it may not always be possible to gather additional data points. Considering such a scenario, we decided to remove these edges in the refining stage to prevent unnecessary data distortion. The refined causal model can be seen in \autoref{fig:compas}(c).     
   %We observe that there exists an undirected edge $Gender \to Race$ and a directed edge from $Age \to Race$. 
   %This sampling bias can be easily verified by seeing their bivariate distribution. The best way to tackle such biases is at the data collection stage. In our case,

   \textbf{Auditing and mitigating bias} As previously observed, African Americans are over-represented in the dataset (60\%) relative to their their population as a whole in Florida (17\%). To further probe the presence of racial bias, we selected \textit{Race} and \textit{Recidivism} from the top panel to see their bivariate distribution in the comparison view (as shown in \autoref{fig:compas}(d)). From the grouped bar chart, we can clearly observe a disparity in the likelihood to recidivate (recommit a crime) between African Americans and Caucasians. As seen in the case of Berkeley's admission dataset, disparity between two groups does not necessarily mean systematic discrimination, so we tried to understand the underpinnings of this disparity via the causal network. 
   %Training a ML model over such a dataset can potentially result in algorithmic bias. We can clearly observe a  that the dataset is biased against African-Americans. 

   We used the find paths functionality to find all paths from \textit{Race} to \textit{Recidivism}. We obtain 3 paths as shown in \autoref{fig:compas}(e). Among the causal paths, we found causal edges such as \textit{Race} $\to$ \textit{Prior offences} and \textit{Race} $\to$ \textit{Charge degree}. From our domain knowledge, we find these causal relationships, inherent in the dataset, to be biased and socially undesirable. Training a ML model over such a dataset can potentially result in replicating, and even amplifying such racial biases. To prevent that, we deleted these causal edges to obtain our debiased causal network (see \autoref{fig:compas}(f)). 
   %To prevent any ML model trained ofrom learning such biases, we choise to  Our objective is to dilute or remove this edges to mitigate bias. Let's take the first path from \textit{Race} $\to$ \textit{Recidivism} via \textit{Prior Offenses}. Here, we have two edges that we can act upon. Based on our domain knowledge, we find that \textit{Prior Offenses} $\to$ \textit{Recidivism} to be socially acceptable while \textit{Race} $\to$ \textit{Prior Offenses} is not. So, we remove this edge. Similarly, we also remove the edge \textit{Race} $\to$ \textit{Charge Degree} to obtain our debiased causal network (see \autoref{fig:compas}(f)). 
   Since both these operations were executed during the debiasing stage, the system generates a new (debiased) dataset which accounts for these changes.   
   Lastly, we evaluated the debiased data in terms of different evaluation metrics. We found that the debiased data performs well for 4 out of the 5 fairness metrics (\autoref {fig:compas}(g)). More specifically, there has been a vast improvement in fairness as captured by Ind. bias, FNR diff and FPR diff metrics. On the flip side, this process has incurred a small loss in accuracy (4\%) and f1 score (5\%), and the data distortion rose to $13\%$. The debiased dataset can be downloaded from the generator panel and can be used in place of the original data for fairer predictions.   
   
\section{Baseline Tool}
In this section, we present some additional details about the baseline tool. The objective of creating the baseline tool is to evaluate our human-in-the-loop AI based approach relative to the automated approach on human-centric measures such as trust, accountability, etc. This is an important contribution as the existing literature on debiasing algorithms has solely focused on different fairness metrics for evaluation. The workflow of the baseline tool mimics that of automated tools such as IBM's AIF 360 and its design matches with our D-BIAS tool. Its visual interface is roughly equivalent to the D-BIAS tool except for the causal network view. The role of the human is quite limited for the baseline tool. The user first encounters the default view and selects the dataset, label variable, nominal variables and the sensitive variable (see \autoref{fig:baseline_stages} (A)). For bias identification, the user simply clicks the ``Check Bias" button to compute and visualize a set of bias and utility metrics as shown in \autoref{fig:baseline_stages}(B). If the bias scores are within an acceptable range, then no further action is needed as there is no significant bias with respect to the sensitive variable. Otherwise, the user can click on the ``Debias \& Evaluate" button to debias the dataset and compute a new set of evaluation metrics. As shown in \autoref{fig:baseline_stages} (C), the user can compare how the debiased dataset performs on different evaluation metrics relative to the baseline (original dataset without the sensitive attribute). 
%In the real world, the user may also opt for a different debiasing algorithm or tweak hyperparameters of the debiasing algorithm if the they are still not satisfied with the results. 
To have a tightly controlled experiment, we ensured that the evaluation metrics (utility metrics, fairness metrics and data distortion) for the baseline tool exactly matches with the peak performance of our D-BIAS tool. In other words, our study compares both tools (representing two different approaches) while controlling for their performance on different evaluation metrics. This helps better evaluate the design of our tool and helps understand if human interaction can enhance trust, accountability, etc. in the context of bias mitigation.   
%This ensures that the difference in human-centric metrics recorded in the user study are indepenet of the evaluation metrics.    

\end{document}